\documentclass{article} 
\usepackage{2027_conference,times}

\usepackage{amsmath,amsfonts,bm}

\def\eqref#1{equation~\ref{#1}}

\def\1{\bm{1}}

\DeclareMathAlphabet{\mathsfit}{\encodingdefault}{\sfdefault}{m}{sl}
\SetMathAlphabet{\mathsfit}{bold}{\encodingdefault}{\sfdefault}{bx}{n}

\usepackage{hyperref}
\usepackage{url}
\usepackage{amsmath}
\usepackage{amssymb}
\usepackage{booktabs}
\usepackage{multirow}
\usepackage{graphicx}
\usepackage[table]{xcolor}
\usepackage{caption}
\usepackage{array}
\usepackage{algorithm}
\usepackage{algpseudocode}
\usepackage{pifont}
\usepackage{makecell}
\usepackage[most]{tcolorbox}
\usepackage{enumitem}
\colorlet{basegray}{black}

\title{GleanVID: Complementary Token Selection for Efficient Video Large Language Models}

\author{%
\textbf{Shuo Yang}$^{1*}$\enspace
\textbf{Changbai Li}$^{1*}$\enspace
\textbf{Rui Tang}$^{1}$\enspace
\textbf{Xinyu Zhao}$^{1}$\enspace
\textbf{Linlin Yang}$^{2}$\enspace
\textbf{Baochang Zhang}$^{1}$\\[3pt]
$^{1}$Beihang University\qquad
$^{2}$Communication University of China
}
\iclrfinalcopy

\begin{document}

\maketitle
\lhead{Preprint.}
{\renewcommand{\thefootnote}{\fnsymbol{footnote}}
\footnotetext[1]{Equal contribution.}}

\begin{abstract}
Video Large Language Models (VideoLLMs) have achieved strong video understanding capabilities but incur substantial inference overhead due to the large number of visual tokens. Existing VideoLLM token compression methods largely rely on selection-independent scoring, overlooking cross-frame complementarity and consequently retaining redundant evidence across frames. Instead, we view video token selection as a progressive evidence accumulation process. It aims to retain visual evidence that is individually informative and collectively complementary under a limited token budget. Building on this insight, we introduce GleanVID, a training-free inference acceleration framework for VideoLLMs. Specifically, GleanVID first allocates the global token budget across frames according to temporal novelty and then selects tokens by jointly considering local representativeness and subspace complementarity, thereby preserving richer and less redundant visual evidence.
Extensive experiments across diverse VideoLLMs and benchmarks demonstrate that GleanVID consistently achieves state-of-the-art performance. Notably, with only 25\% of visual tokens, GleanVID preserves 98.6\% of Qwen3-VL's original performance while reducing its prefill latency by 44.7\%. 
On LLaVA-OV-7B, GleanVID at a 25\% retention ratio even slightly surpasses the original model.

\end{abstract}

\section{Introduction}~\label{sec:Intro}

Video Large Language Models (VideoLLMs)~\citep{DBLP:journals/corr/abs-2501-12386,DBLP:conf/iclr/ChenXLHZLFTYLHY25, yang2025timeexpert,DBLP:journals/corr/abs-2501-13106,DBLP:journals/corr/abs-2412-05271, lin2024video}, which integrate large language models~\citep{DBLP:journals/corr/abs-2303-08774} with visual encoders~\citep{DBLP:journals/corr/abs-2502-14786, DBLP:conf/icml/RadfordKHRGASAM21}, have achieved remarkable performance on complex video understanding tasks such as video question answering and long video summarization. However, the inherently multi-frame nature of video results in a visual token count far exceeding that of a single image: for instance, LLaVA-OneVision~\citep{li2024llavaonevisioneasyvisualtask} processes thousands of visual tokens per video, with the count rising to tens of thousands for long videos. As the computational cost of dense attention grows quadratically with sequence length, while token-wise computation and KV-cache memory also increase with the number of input tokens, long visual sequences constitute a major bottleneck for efficient VideoLLM inference.

\begin{figure}[t]
    \centering
    \includegraphics[width=\linewidth]{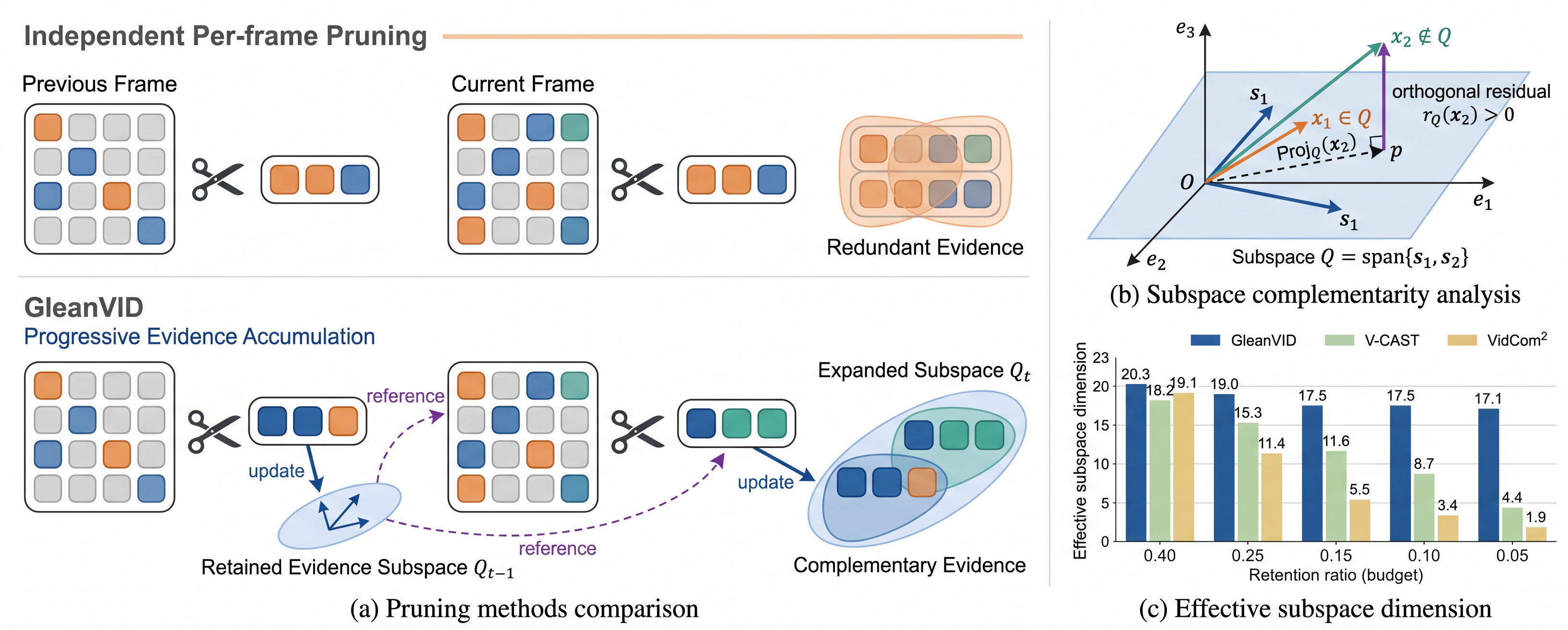}
    \caption{Motivation of GleanVID.
    \textbf{(a)} Existing methods typically prune each frame independently, retaining redundant evidence, whereas GleanVID progressively selects tokens complementary to previously retained evidence.
    \textbf{(b)} Only candidates outside the retained subspace $Q$ (\emph{e.g.}, $x_2$) contribute new information.
    \textbf{(c)} Under tighter budgets, effective dimension collapses in existing methods but remains high for GleanVID.}
    \label{fig:teaser}
    \vspace{-10pt}

\end{figure}

To alleviate this bottleneck, existing VideoLLM compression methods mainly reduce visual redundancy through importance-based pruning~\citep{chen2024image,DBLP:conf/icml/0020FMZ0CGONKZ25,yang2025visionzip,huang2025prunevid}, relation-driven compression~\citep{fu2025framefusion,NEURIPS2025_c573258c,fan2026flashvid,NEURIPS2025_b2e63e36}, or set-aware token selection~\citep{ma2026mmg}. A common practice is to assess token utility primarily using frame-local or token-wise cues, which may repeatedly retain salient content that persists over time (Fig.~\ref{fig:teaser}(a), top). Although recent methods~\citep{ma2026mmg,fan2026flashvid} further exploit cross-frame relations or condition selection on previously retained tokens, they often characterize redundancy through pairwise or token-level interactions, without fully accounting for the information jointly covered by the accumulated evidence. This creates a mismatch with the temporal nature of video: under a limited token budget, effective token selection should progressively accumulate evidence that is individually informative and complementary to what has already been retained across frames (Fig.~\ref{fig:teaser}(a), bottom).

To better understand these limitations, we conduct diagnostic analyses on 100 LongVideoBench videos.
First, across temporal gaps of 1--20 frames, the spatial positions selected by VidCom\textsuperscript{2}~\citep{liu2025video} and V-CAST~\citep{lin2026v} exhibit average IoUs of 0.257 and 0.233, respectively, compared with 0.139 for random selection, suggesting that information-driven criteria tend to revisit persistent spatial regions.
Second, token-level distinctiveness does not imply set-level novelty: a candidate may differ from every selected token yet remain well explained by their joint span (Fig.~\ref{fig:teaser}(b)).
Indeed, tokens selected by a greedy pairwise-diversity strategy retain only 29\% of their feature energy outside the subspace spanned by previous selections.
Third, as the retention ratio decreases from 40\% to 5\%, the effective dimension drops from 18.2 to 4.4 for V-CAST and from 19.1 to 1.9 for VidCom\textsuperscript{2} (Fig.~\ref{fig:teaser}(c)), indicating that the retained evidence collapses into a progressively lower-dimensional subspace.
Together, these observations suggest that existing criteria largely overlook the \emph{joint} diversity of retained evidence, especially under tight budgets.

Building on this analysis, we introduce GleanVID, a training-free inference
acceleration framework that casts video token compression as progressive
cross-frame evidence accumulation under a limited token budget. GleanVID first
employs Temporal Novelty-Guided Budgeting to estimate each frame's temporal
novelty relative to recent visual content and adaptively allocate the
per-frame token budget. Under these budgets, Complementary Token Selection
processes frames sequentially and jointly considers local representativeness
and subspace-level cross-frame complementarity. Local representativeness
favors coherent and distinctive visual content within each frame while
reducing sensitivity to isolated variations, but it does not explicitly
account for evidence retained from earlier frames. We therefore assess
cross-frame complementarity at the subspace level, favoring candidates whose
features are less explainable by combinations of previously retained tokens.

Extensive experiments on three representative VideoLLMs across four
video understanding benchmarks demonstrate that GleanVID achieves a favorable tradeoff between
efficiency and performance across different model architectures and token
budgets. On Qwen3-VL-8B-Instruct with 32 input frames, GleanVID retains
98.6\% of the original performance at a 25\% retention ratio, while reducing
prefill latency by 44.7\%, total latency by 24.1\%, and peak GPU memory by
13.0\%, with a 32.0\% increase in throughput. Moreover, when the retention
ratio decreases from 25\% to 15\%, its relative performance decreases by only 1.6 points, from 98.6\% to 97.0\%, compared with a 1.9-point drop for FlashVID and a 2.8-point drop for
V-CAST, demonstrating greater robustness under tighter token budgets.

Our main contributions are summarized as follows:
\begin{itemize}

\item We identify that common frame-local and pairwise criteria may not fully
capture cross-frame evidence complementarity, and cast video token selection
as progressive evidence accumulation under a limited budget.

\item We propose GleanVID, a training-free VideoLLM token compression framework that allocates per-frame budgets based on temporal novelty and selects tokens by balancing local representativeness with subspace-level cross-frame complementarity.

\item Experiments across multiple VideoLLMs and benchmarks demonstrate a
favorable tradeoff between efficiency and performance, together with stronger
performance retention under tighter token budgets.
\end{itemize}

\section{Related Work}

\textbf{Video Large Language Models.} VideoLLMs integrate visual encoders with large language models for video understanding. Early works such as Video-LLaMA~\citep{zhang2023video} and VideoChat~\citep{li2025videochat} extend image-language models to videos, while LLaVA-OneVision~\citep{li2024llavaonevisioneasyvisualtask}, LLaVA-Video~\citep{DBLP:journals/tmlr/ZhangWLLMLL25}, and Qwen3-VL~\citep{bai2025qwen3vltechnicalreport} further improve performance through large-scale video instruction tuning and spatiotemporal designs such as dynamic-resolution encoding and MRoPE. However, encoding multiple frames yields far more visual tokens than a single image, posing significant challenges to inference efficiency.

\paragraph{Token Compression for VideoLLMs.}
To reduce visual token overhead during VideoLLM inference, recent studies
have proposed a variety of training-free token compression methods, which can
be broadly grouped into importance-guided pruning, relation-driven compression,
and set-aware token selection. FastV~\citep{chen2024image}, SparseVLM~\citep{DBLP:conf/icml/0020FMZ0CGONKZ25}, VisionZip~\citep{yang2025visionzip}, and PruneVid~\citep{huang2025prunevid}
select tokens based on attention, visual saliency, or question relevance.
Relation-driven methods, including FrameFusion~\citep{fu2025framefusion}, DyCoke~\citep{tao2025dycoke}, HoliTom~\citep{NEURIPS2025_c573258c}, FastVID~\citep{NEURIPS2025_b2e63e36}, and
FlashVID~\citep{fan2026flashvid}, exploit spatial or temporal relations for token merging and pruning.
VidCom$^2$~\citep{liu2025video} and V-CAST~\citep{lin2026v} further use inter-frame content variation to allocate
different token budgets across frames. Set-aware methods~\citep{ma2026mmg} condition candidate
utility on previously retained tokens, typically through token-level marginal
gains or pairwise relations. Despite their favorable tradeoffs between
compression ratio and performance, adequately modeling cross-frame evidence
complementarity during token selection remains an important direction for
further investigation.

\section{Methodology}~\label{sec:Method}
\vspace{-20pt}

\subsection{Preliminaries and Overview}
\label{sec:prelim_overview}

Given a video consisting of \(T\) frames, the visual encoder produces a token
tensor
\(\mathbf{X}=[\mathbf{x}_{t,n}]\in\mathbb{R}^{T\times N\times D}\),
where each frame contains \(N\) visual tokens of dimension \(D\). Given a
retention ratio \(\rho\), our goal is to retain a subset
\(\mathcal{S}\) of \(B=\lfloor\rho TN\rfloor\) tokens while minimizing the
degradation in downstream performance. Existing methods typically follow a selection-independent
scoring paradigm. For simplicity, this paradigm can be expressed
as computing a score \(f(\mathbf{x}_i)\) for each token and retaining the
global top-\(B\) tokens:
\begin{equation}
\mathcal{S}
=
\operatorname*{arg\,top\text{-}B}_{i\in\mathcal{I}}
f(\mathbf{x}_i),
\label{eq:static_selection}
\end{equation}
where \(\mathcal{I}\) denotes the index set of all visual tokens.
The scoring function \(f\) may incorporate predefined reference features but
remains unchanged as selection proceeds.

We instead formulate cross-frame token selection as a frame-sequential
decision process:
\begin{equation}
\mathcal{S}_0=\varnothing,\quad
\mathcal{T}_t=
\operatorname*{arg\,top\text{-}k_t}_{i\in\mathcal{I}_t}
g(\mathbf{x}_i;\mathcal{S}_{t-1}),\quad
\mathcal{S}_t=\mathcal{S}_{t-1}\cup\mathcal{T}_t,
\label{eq:sequential_selection}
\end{equation} where \(\mathcal{I}_t\) denotes the token indices in frame \(t\),
\(k_t\) is the token budget allocated to that frame, and
\(\sum_{t=1}^{T}k_t=B\). The scoring function
\(g(\mathbf{x}_i;\mathcal{S}_{t-1})\) evaluates each candidate with respect
to the evidence retained from preceding frames. The selection-independent
formulation in Eq.~\ref{eq:static_selection} is a special case in which
\(g\) does not depend on \(\mathcal{S}_{t-1}\).

\begin{figure}[t]
    \centering
    \includegraphics[width=\textwidth]{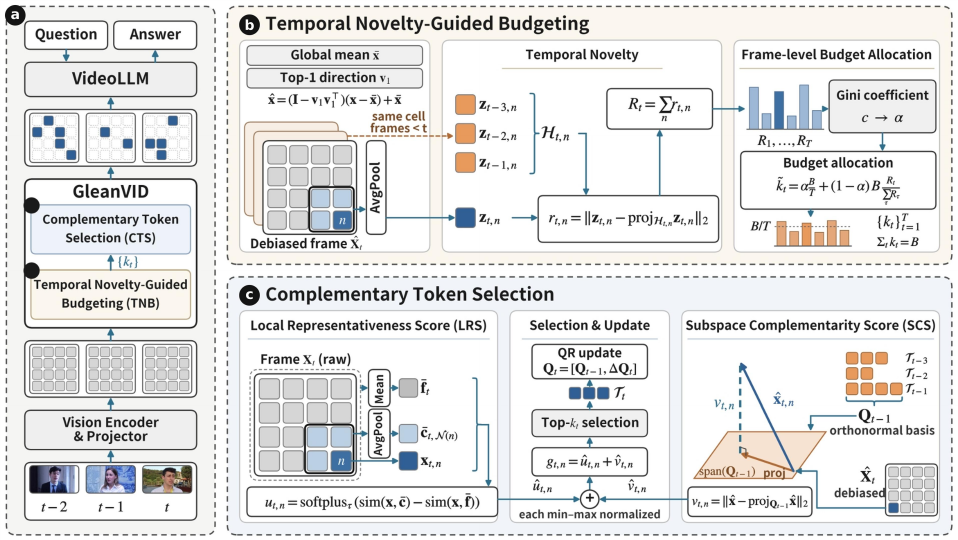}
    \caption{Overview of GleanVID.
    \textbf{(a)} GleanVID compresses the visual tokens of $T$ frames for the VideoLLM.
    \textbf{(b)} TNB measures the novelty of each neighborhood prototype against its history subspace and allocates per-frame budgets $\{k_t\}$ with a Gini-adaptive weight.
    \textbf{(c)} CTS selects the top-$k_t$ tokens by combining LRS with SCS, measured against the basis $\mathbf{Q}_{t-1}$ of retained tokens.}   
    \label{fig:overview}
    \vspace{-2pt}

\end{figure}
Building on Eq.~\ref{eq:sequential_selection}, GleanVID comprises two
components, as illustrated in Fig.~\ref{fig:overview}. Temporal Novelty-Guided Budgeting
(Section~\ref{sec:budgeting}) allocates per-frame token budgets according to
temporal novelty, while Complementary Token Selection
(Section~\ref{sec:selection}) sequentially selects tokens by combining the
Local Representativeness Score (LRS) with the Subspace Complementarity Score
(SCS) measured against the subspace spanned by previously selected tokens.

\subsection{Temporal Novelty-guided Budgeting}
\label{sec:budgeting}

The information content carried by a video at different moments often
varies, motivating per-frame budget allocation. We therefore propose a
dynamic budget allocation strategy guided by content novelty: unlike
segmentation-based merging (e.g., HoliTom, FastVID), our method does not
disrupt tokens' original spatial coordinate structure; and unlike fixed
uniform or random allocation, it adapts each frame's quota to
content-level differences.

\paragraph{Context Debiasing.}
We first center all tokens and remove their shared dominant context direction. Let $\bar{x}$ denote the mean over all tokens, and
$v_1 \in \mathbb{R}^{D}$ the top-1 right singular vector of the centered
feature matrix; the debiased feature of position $n$ in frame $t$, is
\begin{equation}
    \hat{x}_{t,n} = (x_{t,n} - \bar{x}) -
    \big((x_{t,n} - \bar{x})^\top v_1\big) v_1 + \bar{x}.
    \label{eq:debias}
\end{equation}

\paragraph{Temporal Novelty.}
We then quantify each token location's temporal novelty relative to
historical content. 
Directly comparing tokens at identical spatial positions may overestimate novelty when semantically similar content shifts because of object or camera
motion. We therefore average the debiased features within the $2\times2$
pooling cell $\mathcal{N}(n)$ containing position $n$, obtaining the
neighborhood prototype
$z_{t,n} = |\mathcal{N}(n)|^{-1} \sum_{m\in\mathcal{N}(n)}\hat{x}_{t,m}$.
Using the corresponding prototypes from the preceding $h$ frames, we
construct the historical reference set
$H_{t,n} = \{z_{t',n}\}_{t'=\max(1,t-h)}^{t-1}$ and define temporal novelty as
\begin{equation}
    r_{t,n} =
    \begin{cases}
        \lVert z_{t,n} \rVert_2,
        & t = 1, \\[2pt]
        \left\lVert
        z_{t,n}
        -
        \operatorname{proj}_{\operatorname{span}(H_{t,n})}
        z_{t,n}
        \right\rVert_2,
        & t > 1.
    \end{cases}
    \label{eq:novelty}
\end{equation}
Here, $\operatorname{proj}_{\operatorname{span}(H_{t,n})}z_{t,n}$ denotes the orthogonal projection of \(z_{t,n}\) onto the subspace spanned by the historical prototypes in \(H_{t,n}\). A larger $r_{t,n}$ indicates local content that is less predictable
from its historical trajectory and thus more temporally novel.

\paragraph{Frame-level Budget Allocation.}
Summing token-level novelty within each frame, $R_t = \sum_n r_{t,n}$, and
normalizing yields a temporal share $s_t = R_t / \sum_{t'} R_{t'}$; the
budget for each frame mixes uniform and novelty-proportional allocation:
\begin{equation}
    \tilde{k}_t = \alpha \cdot \frac{B}{T} + (1-\alpha) \cdot B \cdot s_t.
    \label{eq:budget}
\end{equation}
Here, \(\alpha\) is the adaptive weight. Videos differ substantially in temporal dynamics: videos with a steady
pace and evenly unfolding information exhibit little variation in
novelty across frames, favoring uniform allocation, whereas videos
containing scene transitions or sudden events concentrate novel
information in a few key frames, favoring novelty-biased allocation. To
let $\alpha$ match a video's own temporal dynamics, we adaptively
determine it via the Gini coefficient of the frame-level novelty
distribution $\{R_t\}_{t=1}^T$,
\begin{equation}
    c = \frac{2\sum_{i=1}^T i \cdot R_{(i)}}{T\sum_{t=1}^T R_t} -
    \frac{T+1}{T},
    \label{eq:gini}
\end{equation}
where $R_{(1)} \le \cdots \le R_{(T)}$ denotes the sorted novelty values.
With its upper and lower bounds,
\begin{equation}
\alpha
=
\alpha_{\max}
-
(\alpha_{\max}-\alpha_{\min})c,
\label{eq:adaptive_alpha}
\end{equation}
so that a more concentrated novelty distribution ($c$ larger) biases the
budget toward novel frames, while a more uniform distribution biases it
toward equal allocation. The final integer budget $k_t$ is obtained via
largest-remainder rounding of $\tilde{k}_t$, ensuring $\sum_t k_t = B$.


\begin{figure*}[t]
    \centering
    \includegraphics[width=\textwidth]{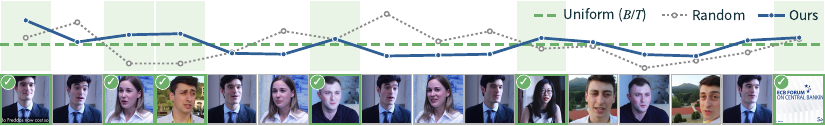}
    \caption{Qualitative comparison of frame-budget allocation under the same global token budget. Curves show the per-frame budget relative to the uniform allocation $B/T$ (dashed line); highlighted frames (\ding{51}) contain newly introduced visual content.}   \label{fig:budget_allocation_case}
    \vspace{-10pt}

\end{figure*}

\subsection{Complementary Token Selection}
\label{sec:selection}

Given the per-frame budgets \(\{k_t\}_{t=1}^{T}\), we next determine which
tokens to retain from each frame. An informative candidate should not only
represent coherent and distinctive local content, but also complement the
evidence accumulated from preceding frames. We therefore evaluate each
candidate using a Local Representativeness Score (LRS) and a Subspace
Complementarity Score (SCS).

\paragraph{Local Representativeness Score (LRS).}
For position $n$ in frame $t$, we compare its raw feature
$x_{t,n}$ against two references: the pooled prototype of its $2\times2$
local neighborhood, $\bar{c}_{t,\mathcal{N}(n)}$, and the global frame
prototype $\bar{f}_t$ (both computed from raw features). Letting
$\mathrm{sim}(\cdot,\cdot)$ denote cosine similarity,
the Local Representativeness Score is defined as
\begin{equation}
    u_{t,n} = \mathrm{softplus}\big(\tau \cdot
    (\mathrm{sim}(x_{t,n}, \bar{c}_{t,\mathcal{N}(n)}) -
    \mathrm{sim}(x_{t,n}, \bar{f}_t))\big) / \tau,
    \label{eq:ncd}
\end{equation}
where $\tau$ controls the sharpness of the mapping. A larger $u_{t,n}$
indicates that the token is more representative of its local
neighborhood, rather than merely aligning with overall frame-level
saliency, favoring locally coherent and
distinctive visual content while reducing sensitivity to isolated variations.

\paragraph{Subspace Complementarity Score (SCS).}
Local representativeness alone does not account for evidence already retained
from preceding frames. We therefore maintain an orthonormal basis
\(\mathbf{Q}_{t-1}\), initialized as empty ($Q_0=\emptyset$), spanning the debiased features of previously selected
tokens, which represents the visual evidence accumulated before frame \(t\).
The Subspace Complementarity Score of a candidate is defined by its
orthogonal projection residual:
\begin{equation}
    v_{t,n} = \big\| \hat{x}_{t,n} -
    \operatorname*{proj}_{Q_{t-1}} \hat{x}_{t,n} \big\|.
    \label{eq:scr}
\end{equation}
A larger \(v_{t,n}\) indicates that the candidate contains a larger component
not explained by the previously retained evidence and therefore provides
greater cross-frame complementarity. Unlike pairwise similarity, SCS
evaluates each candidate against the subspace jointly spanned by all
previously selected tokens, allowing it to detect redundancy collectively
covered by multiple pieces of evidence.

Within each frame, we independently min-max normalize
\(\{u_{t,n}\}_{n=1}^{N}\) and \(\{v_{t,n}\}_{n=1}^{N}\) to \([0,1]\),
obtaining \(\hat{u}_{t,n}\) and \(\hat{v}_{t,n}\). We instantiate the
selection score in Eq.~\ref{eq:sequential_selection} as
\begin{equation}
g(\mathbf{x}_{t,n};\mathcal{S}_{t-1})
=
\hat{u}_{t,n}+\hat{v}_{t,n}.
\label{eq:joint_score}
\end{equation}
We retain the top-\(k_t\) candidates according to
Eq.~\ref{eq:joint_score} to form \(\mathcal{T}_t\). Their orthogonal
components relative to \(\mathbf{Q}_{t-1}\) are then incorporated into the
basis to obtain \(\mathbf{Q}_t\) for the subsequent frame. This
frame-sequential procedure progressively accumulates locally representative
and cross-frame complementary visual evidence under the global token budget.

\section{Experiments}~\label{sec:Experiment}
\vspace{-20pt}

\subsection{Experimental Settings}
\label{sec:exp_settings}

\newcommand{\methodtag}[1]{\,\mbox{\scriptsize\textnormal{(#1)}}}
\definecolor{oursbg}{RGB}{234,239,244}
\begin{table*}[t]
    \centering
    \captionsetup{position=top,skip=3pt}
    \caption{Performance comparison with existing baselines on Qwen3-VL-8B-Instruct across multiple benchmarks and input settings. ``Average'' denotes the mean performance across benchmarks.}
    \label{tab:qwen3vl-8b}
    \setlength{\tabcolsep}{4.2pt}
    \renewcommand{\arraystretch}{1.06}
    \resizebox{\textwidth}{!}{%
    \begin{tabular}{lcccccccccc}
        \toprule
        \multirow{2}{*}{\textbf{Method}} &
        \multirow{2}{*}{\textbf{MVBench}} &
        \multirow{2}{*}{\shortstack{\textbf{LongVideo}\\\textbf{Bench}}} &
        \multirow{2}{*}{\textbf{MLVU}} &
        \multicolumn{4}{c}{\textbf{VideoMME}} &
        \multicolumn{2}{c}{\textbf{Average}} \\
        \cmidrule(lr){5-8}\cmidrule(lr){9-10}
        & & & & \textbf{Overall} & \textbf{Short} & \textbf{Medium} & \textbf{Long} & \textbf{Score} & \textbf{\%} \\
        \midrule

        \multicolumn{10}{c}{\textit{Max Input Frames = 64}} \\
        \midrule

        \rowcolor{gray!10}
        \textcolor{basegray}{Qwen3-VL-8B-Instruct} &
        \textcolor{basegray}{69.2} & \textcolor{basegray}{62.8} & \textcolor{basegray}{68.9} &
        \textcolor{basegray}{66.9} & \textcolor{basegray}{78.8} & \textcolor{basegray}{66.2} &
        \textcolor{basegray}{55.8} & \textcolor{basegray}{67.0} & \textcolor{basegray}{100.0} \\
        \midrule

        \multicolumn{10}{l}{\textit{Retention Ratio = 25\%}} \\
        VisionZip\methodtag{CVPR'25} & 64.8 & 58.9 & 63.4 & 63.1 & 73.7 & 60.3 & 55.4 & 62.6 & 93.4 \\
        VidCom\textsuperscript{2}\methodtag{EMNLP'25} & 67.5 & 59.6 & 64.0 & 64.9 & 75.4 & \textbf{63.4} & \underline{55.9} & 64.0 & 95.5 \\
        FastVID\methodtag{NeurIPS'25} & 66.8 & 60.4 & 65.0 & 62.3 & 73.9 & 60.7 & 52.2 & 63.6 & 94.9 \\
        HoliTom\methodtag{NeurIPS'25} & 64.8 & 58.9 & 63.4 & 62.7 & 74.4 & 60.6 & 53.2 & 62.5 & 93.3 \\
        FlashVID\methodtag{ICLR'26} & OOM & OOM & OOM & OOM & OOM & OOM & OOM & -- & -- \\
        V-CAST\methodtag{2026'03} & \underline{67.7} & \textbf{61.2} & \underline{65.3} & \underline{65.1} & \underline{76.8} & 63.2 & 55.3 & \underline{64.8} & \underline{96.7} \\
        \rowcolor{oursbg}
        \textbf{GleanVID}\methodtag{Ours} & \textbf{68.0} & \underline{61.1} & \textbf{65.8} & \textbf{66.0} & \textbf{77.8} & \underline{63.3} & \textbf{56.9} & \textbf{65.2} & \textbf{97.3} \\
        \midrule

        \multicolumn{10}{l}{\textit{Retention Ratio = 15\%}} \\
        VisionZip\methodtag{CVPR'25} & 62.0 & 57.7 & 61.4 & 60.4 & 70.3 & 58.4 & 52.6 & 60.4 & 90.1 \\
        VidCom\textsuperscript{2}\methodtag{EMNLP'25} & 64.3 & 57.4 & 60.3 & 62.9 & 72.6 & 61.3 & \underline{54.7} & 61.2 & 91.3 \\
        FastVID\methodtag{NeurIPS'25} & 64.9 & \underline{59.7} & 63.3 & 60.5 & 72.2 & 57.2 & 52.1 & 62.1 & 92.7 \\
        HoliTom\methodtag{NeurIPS'25} & 62.7 & 58.3 & 61.8 & 60.9 & 72.3 & 58.2 & 52.2 & 60.9 & 90.9 \\
        FlashVID\methodtag{ICLR'26} & OOM & OOM & OOM & OOM & OOM & OOM & OOM & -- & -- \\
        V-CAST\methodtag{2026'03} & \underline{66.0} & 59.6 & \underline{64.4} & \underline{63.8} & \underline{76.3} & \underline{61.3} & 53.7 & \underline{63.5} & \underline{94.8} \\
        \rowcolor{oursbg}
        \textbf{GleanVID}\methodtag{Ours} & \textbf{67.1} & \textbf{60.0} & \textbf{65.2} & \textbf{64.8} & \textbf{76.8} & \textbf{62.4} & \textbf{55.2} & \textbf{64.3} & \textbf{96.0} \\
        \midrule

        \multicolumn{10}{c}{\textit{Max Input Frames = 32}} \\
        \midrule

        \rowcolor{gray!10}
        \textcolor{basegray}{Qwen3-VL-8B-Instruct} &
        \textcolor{basegray}{68.6} & \textcolor{basegray}{60.3} & \textcolor{basegray}{63.5} &
        \textcolor{basegray}{64.5} & \textcolor{basegray}{76.0} & \textcolor{basegray}{60.4} &
        \textcolor{basegray}{57.0} & \textcolor{basegray}{64.2} & \textcolor{basegray}{100.0} \\
        \midrule

        \multicolumn{10}{l}{\textit{Retention Ratio = 25\%}} \\
        VisionZip\methodtag{CVPR'25} & 62.2 & 56.7 & 60.8 & 60.1 & 69.6 & 56.7 & 54.2 & 60.0 & 93.5 \\
        VidCom\textsuperscript{2}\methodtag{EMNLP'25} & 67.0 & 58.0 & 60.6 & 62.4 & 72.1 & 59.1 & \textbf{56.1} & 62.0 & 96.6 \\
        FastVID\methodtag{NeurIPS'25} & 67.3 & 58.7 & 60.7 & 60.5 & 72.0 & 56.8 & 52.7 & 61.8 & 96.3 \\
        HoliTom\methodtag{NeurIPS'25} & 63.0 & 56.8 & 61.2 & 59.7 & 71.4 & 54.6 & 53.1 & 60.2 & 93.8 \\
        FlashVID\methodtag{ICLR'26} & 67.5 & \underline{58.8} & 61.7 & 62.3 & 74.4 & 58.7 & 53.9 & 62.6 & 97.5 \\
        V-CAST\methodtag{2026'03} & \underline{67.5} & 58.1 & \textbf{62.7} & \underline{63.5} & \underline{74.4} & \underline{60.2} & \underline{55.8} & \underline{62.9} & \underline{98.0} \\
        \rowcolor{oursbg}
        \textbf{GleanVID}\methodtag{Ours} & \textbf{68.0} & \textbf{59.7} & \underline{61.8} & \textbf{63.6} & \textbf{74.6} & \textbf{60.7} & 55.7 & \textbf{63.3} & \textbf{98.6} \\
        \midrule

        \multicolumn{10}{l}{\textit{Retention Ratio = 15\%}} \\
        VisionZip\methodtag{CVPR'25} & 60.1 & 56.2 & 60.0 & 58.2 & 66.9 & 54.9 & 52.9 & 58.6 & 91.3 \\
        VidCom\textsuperscript{2}\methodtag{EMNLP'25} & 64.2 & 56.0 & 57.7 & 59.6 & 68.9 & 56.7 & 53.3 & 59.4 & 92.5 \\
        FastVID\methodtag{NeurIPS'25} & \underline{66.1} & 57.1 & 59.0 & 58.3 & 69.6 & 54.3 & 51.1 & 60.1 & 93.6 \\
        HoliTom\methodtag{NeurIPS'25} & 60.0 & 55.8 & 59.6 & 58.3 & 68.9 & 54.7 & 51.7 & 58.4 & 91.0 \\
        FlashVID\methodtag{ICLR'26} & \textbf{66.5} & \underline{57.8} & 60.0 & 61.4 & \underline{72.4} & 57.6 & 54.1 & \underline{61.4} & \underline{95.6} \\
        V-CAST\methodtag{2026'03} & 64.8 & 57.6 & \underline{60.0} & \underline{62.0} & 71.9 & \underline{58.3} & \textbf{55.9} & 61.1 & 95.2 \\
        \rowcolor{oursbg}
        \textbf{GleanVID}\methodtag{Ours} & 66.0 & \textbf{59.0} & \textbf{61.4} & \textbf{62.6} & \textbf{73.6} & \textbf{59.6} & \underline{54.6} & \textbf{62.3} & \textbf{97.0} \\
        \bottomrule
    \end{tabular}%
    }
    \vspace{-10pt}

\end{table*}

\providecommand{\methodtag}[1]{\,\mbox{\scriptsize\textnormal{(#1)}}}
\providecolor{oursbg}{RGB}{250,244,246}

\begin{table*}[t]
    \centering
    \captionsetup{position=top,skip=3pt}
    \caption{Performance comparison with existing baselines on LLaVA-OV-7B across different benchmarks. We use the default 32-frame input setting.}
    \label{tab:llava-ov}
    \setlength{\tabcolsep}{4.2pt}
    \renewcommand{\arraystretch}{1.06}
    \resizebox{\textwidth}{!}{%
    \begin{tabular}{lcccccccccc}
        \toprule
        \multirow{2}{*}{\textbf{Method}} &
        \multirow{2}{*}{\textbf{MVBench}} &
        \multirow{2}{*}{\shortstack{\textbf{LongVideo}\\\textbf{Bench}}} &
        \multirow{2}{*}{\textbf{MLVU}} &
        \multicolumn{4}{c}{\textbf{VideoMME}} &
        \multicolumn{2}{c}{\textbf{Average}} \\
        \cmidrule(lr){5-8}\cmidrule(lr){9-10}
        & & & & \textbf{Overall} & \textbf{Short} & \textbf{Medium} & \textbf{Long} & \textbf{Score} & \textbf{\%} \\
        \midrule

        \rowcolor{gray!10}
        \textcolor{basegray}{LLaVA-OV-7B} &
        \textcolor{basegray}{58.3} & \textcolor{basegray}{56.6} & \textcolor{basegray}{63.1} &
        \textcolor{basegray}{58.4} & \textcolor{basegray}{69.9} & \textcolor{basegray}{56.7} &
        \textcolor{basegray}{48.8} & \textcolor{basegray}{59.1} & \textcolor{basegray}{100.0} \\
        \midrule
        \multicolumn{10}{l}{\textit{Retention Ratio = 25\%}} \\
        FastV\methodtag{ECCV'24} & 55.5 & 53.3 & 59.6 & 55.3 & 65.0 & 53.8 & 47.0 & 55.9 & 94.6 \\
        SparseVLM\methodtag{ICML'25} & 56.4 & 53.9 & 60.7 & 57.3 & 68.4 & 55.2 & 48.1 & 57.1 & 96.6 \\
        VisionZip\methodtag{CVPR'25} & 56.9 & 56.0 & 62.9 & 58.0 & 68.9 & \underline{57.4} & 47.6 & \underline{58.5} & \underline{99.0} \\
        PruneVid\methodtag{ACL'25} & 55.7 & 55.1 & \textbf{63.4} & 57.0 & 68.8 & 54.4 & 47.7 & 57.8 & 97.8 \\
        FrameFusion\methodtag{ICCV'25} & 56.0 & 54.8 & 61.7 & 57.5 & 68.2 & 55.7 & 48.6 & 57.5 & 97.3 \\
        FastVID\methodtag{NeurIPS'25} & 56.5 & \underline{56.3} & 60.9 & 58.3 & 69.4 & \textbf{58.2} & 47.2 & 58.0 & 98.1 \\
        VidCom\textsuperscript{2}\methodtag{EMNLP'25} & \underline{57.0} & 55.4 & 62.8 & \underline{58.4} & 69.3 & 56.3 & \underline{49.4} & 58.4 & 98.8 \\
        V-CAST\methodtag{2026'03} & 56.1 & 55.9 & 62.9 & 58.2 & \underline{69.9} & 56.0 & 48.7 & 58.3 & 98.6 \\
        \rowcolor{oursbg}
        \textbf{GleanVID}\methodtag{Ours} & \textbf{57.1} & \textbf{57.3} & \underline{63.0} & \textbf{59.7} & \textbf{72.3} & 56.8 & \textbf{50.0} & \textbf{59.3} & \textbf{100.3} \\
        \bottomrule
    \end{tabular}%
    }
\end{table*}

\providecommand{\methodtag}[1]{\,\mbox{\scriptsize\textnormal{(#1)}}}
\providecolor{oursbg}{RGB}{250,244,246}

\begin{table*}[t]
    \centering
    \captionsetup{position=top,skip=3pt}
    \caption{Performance comparison with existing baselines on LLaVA-Video-7B across different benchmarks. We use the default 64-frame input setting.}
    \label{tab:llava-video}
    \setlength{\tabcolsep}{4.2pt}
    \renewcommand{\arraystretch}{1.06}
    \resizebox{\textwidth}{!}{%
    \begin{tabular}{lcccccccccc}
        \toprule
        \multirow{2}{*}{\textbf{Method}} &
        \multirow{2}{*}{\textbf{MVBench}} &
        \multirow{2}{*}{\shortstack{\textbf{LongVideo}\\\textbf{Bench}}} &
        \multirow{2}{*}{\textbf{MLVU}} &
        \multicolumn{4}{c}{\textbf{VideoMME}} &
        \multicolumn{2}{c}{\textbf{Average}} \\
        \cmidrule(lr){5-8}\cmidrule(lr){9-10}
        & & & & \textbf{Overall} & \textbf{Short} & \textbf{Medium} & \textbf{Long} & \textbf{Score} & \textbf{\%} \\
        \midrule

        \rowcolor{gray!10}
        \textcolor{basegray}{LLaVA-Video-7B} &
        \textcolor{basegray}{60.4} & \textcolor{basegray}{58.9} & \textcolor{basegray}{67.3} &
        \textcolor{basegray}{64.4} & \textcolor{basegray}{77.3} & \textcolor{basegray}{62.4} &
        \textcolor{basegray}{53.4} & \textcolor{basegray}{62.8} & \textcolor{basegray}{100.0} \\
        \midrule
        \multicolumn{10}{l}{\textit{Retention Ratio = 25\%}} \\
        FastV\methodtag{ECCV'24} & 52.1 & 54.8 & 57.8 & 58.6 & 68.7 & 58.4 & 48.7 & 55.8 & 88.9 \\
        SparseVLM\methodtag{ICML'25} & 55.4 & 54.2 & 58.9 & 60.1 & 71.1 & 59.1 & 50.1 & 57.2 & 91.1 \\
        VisionZip\methodtag{CVPR'25} & 57.9 & 56.3 & \textbf{62.6} & 62.5 & 73.6 & 62.3 & \underline{51.9} & 59.8 & 95.2 \\
        HoliTom\methodtag{NeurIPS'25} & \underline{58.4} & 57.1 & 60.5 & \textbf{63.0} & \textbf{74.6} & \underline{62.3} & \textbf{52.1} & \underline{59.8} & \underline{95.2} \\
        VidCom\textsuperscript{2}\methodtag{EMNLP'25} & 57.0 & \underline{57.1} & 58.7 & 61.7 & 73.0 & 61.7 & 50.0 & 58.6 & 93.3 \\
        V-CAST\methodtag{2026'03} & 57.8 & 56.8 & 58.2 & 61.5 & 72.2 & 60.9 & 51.3 & 58.6 & 93.3 \\
        \rowcolor{oursbg}
        \textbf{GleanVID}\methodtag{Ours} & \textbf{59.6} & \textbf{57.2} & \underline{60.5} & \underline{62.6} & \underline{73.7} & \textbf{62.4} & 51.6 & \textbf{60.0} & \textbf{95.5} \\
        \bottomrule
    \end{tabular}%
    }
    \vspace{-10pt}

\end{table*}

\paragraph{Benchmarks.}
We evaluate GleanVID on four widely-used video understanding benchmarks: MVBench~\citep{DBLP:conf/cvpr/0002WH00LWX0L0024}, LongVideoBench~\citep{wu2024longvideobench}, MLVU~\citep{Zhou2024MLVUAC}, and VideoMME~\citep{DBLP:conf/cvpr/FuDLLRZWZSZCLLZ25}. These benchmarks span a diverse range of scenarios, from short to long videos and from single-scene understanding to complex multi-task reasoning, enabling a comprehensive evaluation of both the effectiveness and generalization of our method.

\paragraph{Implementation Details.}
We evaluate GleanVID on three representative VideoLLMs with diverse architectures, LLaVA-OneVision~\citep{li2024llavaonevisioneasyvisualtask}, LLaVA-Video~\citep{DBLP:journals/tmlr/ZhangWLLMLL25}, and Qwen3-VL~\citep{bai2025qwen3vltechnicalreport}, to assess its generality. LLaVA-OneVision and LLaVA-Video sample 32 and 64 frames by default, respectively. We evaluate all methods under two retention ratios, 25\% and 15\%, following the same evaluation protocol for a fair comparison. All experiments are conducted on NVIDIA H800-80G GPUs using LMMs-Eval~\citep{Zhang_2025}. Additional hyperparameter settings are provided in the Appendix~\ref{app:setup}.

\subsection{Main Results}
\label{sec:main_results}

\paragraph{Results on Qwen3-VL-8B-Instruct.}
Table~\ref{tab:qwen3vl-8b} compares GleanVID with representative token compression methods under 32- and 64-frame input settings. GleanVID achieves the highest average relative performance across all settings. For example, under the 32-frame setting with a 15\% retention ratio, GleanVID preserves
97.0\% of the original performance, outperforming the strongest baseline, FlashVID (95.6\%), and the image-oriented VisionZip (91.3\%). As the retention ratio decreases from 25\% to 15\%, GleanVID's relative performance drops by only 1.6 points, whereas VidCom$^2$ and HoliTom drop by 4.1 and 2.8 points, respectively. This suggests that progressively accumulating complementary cross-frame evidence is particularly beneficial under tighter token budgets.
Moreover, FlashVID runs out of memory (OOM) under the 64-frame setting, as its attention-based selection requires materialized \([\mathrm{CLS}]\) attention scores, limiting its compatibility with memory-efficient operators such as FlashAttention. In contrast, GleanVID requires no access to attention weights and runs without OOM under the same setting.

\paragraph{Results on LLaVA-OneVision-7B.}
We further extend our evaluation to the LLaVA series of VideoLLMs to
verify the generalization of our method across different model
architectures. Table~\ref{tab:llava-ov} reports results under the 25\% retention ratio: GleanVID achieves 100.3\% relative performance,
surpassing VisionZip (99.0\%) and VidCom\textsuperscript{2} (98.8\%), and slightly outperforms the vanilla model with full token input. This result suggests that the original visual token
sequence may contain some cross-frame redundancy, and that GleanVID's temporal novelty-guided budgeting and complementary token selection may help reduce its potential interference with reasoning, demonstrating the effectiveness of GleanVID.

\vspace{-8pt}

\paragraph{Results on LLaVA-Video-7B.}
We further evaluate GleanVID on LLaVA-Video. Unlike LLaVA-OneVision,
LLaVA-Video samples 64 frames by default and introduces newline tokens to explicitly encode spatiotemporal positional information. Table~\ref{tab:llava-video} reports results under the 25\% retention ratio: despite this architectural difference, GleanVID again achieves the best performance (95.5\% relative performance), surpassing the second-best HoliTom (95.2\%). These results show that GleanVID remains effective under a different
visual-token organization, supporting its applicability across VideoLLM architectures.

\textbf{Results under a Fixed Token Budget.} Keeping the post-compression visual-token count equal to that of the uncompressed 16-frame baseline, GleanVID scales the input to 80 and 160 frames and achieves 110.4\% and 112.4\% relative performance, respectively, effectively converting token compression into broader temporal coverage (Appendix~\ref{app:fixed_budget}).

\providecommand{\methodtag}[1]{\,\mbox{\scriptsize\textnormal{(#1)}}}

\providecolor{oursbg}{RGB}{250,244,246}
\providecolor{improve}{RGB}{45,135,84}
\providecolor{degrade}{RGB}{204,58,58}

\providecommand{\effdown}[1]{\,{\scriptsize
\textcolor{improve}{($\downarrow$#1)}}}

\providecommand{\effup}[1]{\,{\scriptsize
\textcolor{degrade}{($\uparrow$#1)}}}

\providecommand{\effspeed}[1]{\,{\scriptsize
\textcolor{improve}{(#1$\times$)}}}

\providecommand{\perfdrop}[1]{\,{\scriptsize
\textcolor{degrade}{($\downarrow$#1)}}}

\begin{table*}[t]
    \centering
    \captionsetup{position=top,skip=6pt}
    \caption{
        Efficiency comparison on VideoMME using Qwen3-VL-8B-Instruct
        with a retention ratio of $\rho=25\%$ and a maximum of 32 input
        frames. ``Prefill Latency'' denotes the prompt-to-first-token time;
        ``LLM Generation Latency'' denotes the first-to-last-token decoding
        time; ``Total Latency'' denotes the end-to-end wall-clock time in
        our experimental setup; and ``Throughput'' is measured in items per
        second.
    }
    \label{tab:efficiency}

    \setlength{\tabcolsep}{4.0pt}
    \renewcommand{\arraystretch}{1.10}

    \resizebox{\textwidth}{!}{%
    \begin{tabular}{lcccccc}
        \toprule
        \textbf{Method} &
        \shortstack{\textbf{Prefill Latency} $\downarrow$\\
                    \textbf{(s)}} &
        \shortstack{\textbf{LLM Generation} $\downarrow$\\
                    \textbf{Latency (s)}} &
        \shortstack{\textbf{Total Latency} $\downarrow$\\
                    \textbf{(s)}} &
        \shortstack{\textbf{GPU Peak Memory} $\downarrow$\\
                    \textbf{(MB)}} &
        \shortstack{\textbf{Throughput} $\uparrow$\\
                    \textbf{(items/s)}} &
        \textbf{Performance} $\uparrow$ \\
        \midrule

        \rowcolor{gray!10}
        \textcolor{basegray}{Qwen3-VL-8B-Instruct} &
        \textcolor{basegray}{239.9} &
        \textcolor{basegray}{280.2} &
        \textcolor{basegray}{1369.5} &
        \textcolor{basegray}{22478.0} &
        \textcolor{basegray}{1.97} &
        \textcolor{basegray}{64.5} \\
        \midrule

        VidCom\textsuperscript{2}\methodtag{EMNLP'25} &
        \textbf{119.0}\effdown{50.4\%} &
        \textbf{154.1}\effdown{45.0\%} &
        \textbf{1033.6}\effdown{24.5\%} &
        \textbf{19547.5}\effdown{13.0\%} &
        \textbf{2.61}\effspeed{1.32} &
        62.4\perfdrop{2.1} \\

        HoliTom\methodtag{NeurIPS'25} &
        134.3\effdown{44.0\%} &
        169.6\effdown{39.5\%} &
        1063.2\effdown{22.4\%} &
        \underline{19974.7}\effdown{11.1\%} &
        2.54\effspeed{1.29} &
        60.2\perfdrop{4.3} \\

        FlashVID\methodtag{ICLR'26} &
        145.0\effdown{39.6\%} &
        179.0\effdown{36.1\%} &
        1132.2\effdown{17.3\%} &
        40199.2\effup{78.8\%} &
        2.38\effspeed{1.21} &
        62.3\perfdrop{2.2} \\

        V-CAST\methodtag{2026'03} &
        \underline{121.2}\effdown{49.5\%} &
        \underline{159.9}\effdown{42.9\%} &
        \underline{1039.7}\effdown{24.1\%} &
        \textbf{19547.5}\effdown{13.0\%} &
        \underline{2.60}\effspeed{1.32} &
        \underline{63.5}\perfdrop{1.0} \\

        \rowcolor{oursbg}
        \textbf{GleanVID}\methodtag{Ours} &
        132.6\effdown{44.7\%} &
        165.7\effdown{40.9\%} &
        1040.1\effdown{24.1\%} &
        \textbf{19547.5}\effdown{13.0\%} &
        \underline{2.60}\effspeed{1.32} &
        \textbf{63.6}\perfdrop{0.9} \\

        \bottomrule
    \end{tabular}%
    }
    \vspace{-8pt}
    
\end{table*}
\subsection{Ablation Study}
\label{sec:ablation}

\definecolor{dropred}{RGB}{190,45,45}

\newcommand{\drop}[1]{%
    \hspace{0.8pt}%
    \raisebox{0.1ex}{%
        \scalebox{0.85}{%
            \textcolor{dropred}{\tiny($\downarrow$#1)}%
        }%
    }%
}
\begin{table}[t]
    \centering

    \begin{minipage}[t]{0.485\textwidth}
        \vspace{0pt}

        \begin{minipage}[t][3.8\baselineskip][t]{\linewidth}
            \captionof{table}{
                Component and token-selection ablations.
                Relative performance is reported in \% and its drop in points.
            }
            \label{tab:component_signal_ablation}
        \end{minipage}

        \centering
        \scriptsize
        \setlength{\tabcolsep}{0pt}
        \renewcommand{\arraystretch}{1.20}

        \begin{tabular}{
            @{}
            >{\raggedright\arraybackslash}m{0.250\linewidth}
            @{\hspace{0.5pt}}
            >{\centering\arraybackslash}m{0.140\linewidth}
            @{\hspace{1.5pt}}
            >{\centering\arraybackslash}m{0.090\linewidth}
            @{\hspace{1.5pt}}
            >{\centering\arraybackslash}m{0.110\linewidth}
            @{\hspace{1.5pt}}
            >{\centering\arraybackslash}m{0.160\linewidth}
            @{\hspace{1.5pt}}
            >{\centering\arraybackslash}m{0.200\linewidth}
            @{}
        }
            \toprule

            \textbf{Variant} &
            \textbf{MVBench} &
            \textbf{LVB} &
            \textbf{MLVU} &
            \textbf{VideoMME} &
            \shortstack[c]{\textbf{Rel. Perf.}\\\textbf{(\%)}} \\

            \midrule

            \multicolumn{6}{@{}l}{
                \textit{(a) Model Components}
            } \\
            \addlinespace[1.5pt]

            w/o TNB &
            65.5 &
            58.0 &
            60.4 &
            61.2 &
            \mbox{95.4\drop{1.6}} \\

            w/o CTS &
            65.8 &
            56.3 &
            59.5 &
            59.5 &
            \mbox{93.9\drop{3.1}} \\

            \textbf{GleanVID} &
            \textbf{66.0} &
            \textbf{59.0} &
            \textbf{61.4} &
            \textbf{62.6} &
            \textbf{97.0} \\

            \midrule

            \multicolumn{6}{@{}l}{
                \textit{(b) Token-Selection Signals}
            } \\
            \addlinespace[1.5pt]

            LRS only &
            64.8 &
            55.5 &
            59.5 &
            62.0 &
            \mbox{94.2\drop{2.8}} \\

            SCS only &
            \textbf{66.2} &
            58.7 &
            61.0 &
            60.1 &
            \mbox{95.8\drop{1.2}} \\

            LRS + SCS &
            66.0 &
            \textbf{59.0} &
            \textbf{61.4} &
            \textbf{62.6} &
            \textbf{97.0} \\

            \bottomrule
        \end{tabular}
    \end{minipage}
    \hfill
    \begin{minipage}[t]{0.485\textwidth}
        \vspace{0pt}

        \begin{minipage}[t][3.8\baselineskip][t]{\linewidth}
            \captionof{table}{
                Frame-budget and complementarity ablations.
                Relative performance is reported in \% and its drop in points.
            }
            \label{tab:budget_complementarity_ablation}
        \end{minipage}

        \centering
        \scriptsize
        \setlength{\tabcolsep}{0pt}
        \renewcommand{\arraystretch}{1.20}

        \begin{tabular}{
            @{}
            >{\raggedright\arraybackslash}m{0.250\linewidth}
            @{\hspace{0.5pt}}
            >{\centering\arraybackslash}m{0.140\linewidth}
            @{\hspace{1.5pt}}
            >{\centering\arraybackslash}m{0.090\linewidth}
            @{\hspace{1.5pt}}
            >{\centering\arraybackslash}m{0.110\linewidth}
            @{\hspace{1.5pt}}
            >{\centering\arraybackslash}m{0.160\linewidth}
            @{\hspace{1.5pt}}
            >{\centering\arraybackslash}m{0.200\linewidth}
            @{}
        }
            \toprule

            \textbf{Variant} &
            \textbf{MVBench} &
            \textbf{LVB} &
            \textbf{MLVU} &
            \textbf{VideoMME} &
            \shortstack[c]{\textbf{Rel. Perf.}\\\textbf{(\%)}} \\

            \midrule

            \multicolumn{6}{@{}l}{
                \textit{(a) Frame-Budget Allocation}
            } \\
            \addlinespace[1.5pt]

            Random &
            65.5 &
            58.0 &
            60.4 &
            61.2 &
            \mbox{95.4\drop{1.6}} \\

            Uniform &
            64.8 &
            58.3 &
            60.6 &
            62.1 &
            \mbox{95.8\drop{1.2}} \\

            Fixed ($\alpha{=}0.5$) &
            \textbf{66.1} &
            58.5 &
            61.1 &
            62.4 &
            \mbox{96.6\drop{0.4}} \\

            Adaptive &
            66.0 &
            \textbf{59.0} &
            \textbf{61.4} &
            \textbf{62.6} &
            \textbf{97.0} \\

            \midrule

            \multicolumn{6}{@{}l}{
                \textit{(b) Complementarity Measure}
            } \\
            \addlinespace[1.5pt]

            \mbox{Pairwise Distance} &
            65.6 &
            58.7 &
            60.1 &
            61.8 &
            \mbox{95.9\drop{1.1}} \\

            SCS &
            \textbf{66.0} &
            \textbf{59.0} &
            \textbf{61.4} &
            \textbf{62.6} &
            \textbf{97.0} \\

            \bottomrule
        \end{tabular}
    \end{minipage}

\end{table}

We conduct ablation studies on Qwen3-VL-8B-Instruct with 32-frame input and
a 15\% retention ratio, examining model components, frame-budget allocation
strategies, token-selection signals, and complementarity modeling. Additional
hyperparameter sensitivity analyses are provided in
Appendix~\ref{app:additional_ablation}.

\vspace{-5pt}

\paragraph{Ablation study on Model Components.}
Table~\ref{tab:component_signal_ablation}(a) verifies the contributions of TNB and CTS. Removing TNB (degenerating to random budget
allocation) leads to a drop of 1.6 points in relative performance, showing
that, compared to random allocation, capturing the temporal distribution
of video information via content-level novelty provides a more
reasonable budget basis for each frame. Removing CTS (degenerating to
random selection) causes a larger drop of 3.1 points. Taken together, these results demonstrate that GleanVID benefits from
coordinated decisions at two complementary granularities: TNB adapts the
frame-level token budget to the temporal distribution of video information,
while CTS progressively constructs a compact and informative representation
under the resulting budgets.

\vspace{-5pt}

\paragraph{Ablation study on Frame-Budget Allocation.}
Table~\ref{tab:budget_complementarity_ablation}(a) compares different budget allocation strategies within TNB. Both random and uniform allocation disregard the temporal variation intrinsic to video content, resulting in relatively
lower average performance. Figure~\ref{fig:budget_allocation_case}
qualitatively illustrates their different allocation behaviors. Incorporating temporal novelty with a fixed mixing weight of $\alpha=0.5$ brings some improvement but behaves inconsistently across benchmarks. In contrast, our adaptive $\alpha$ strategy, which dynamically adjusts the weight according to the concentration of the frame-level novelty distribution, achieves the best relative performance across benchmarks.

\paragraph{Ablation study on Token Selection.}
Table~\ref{tab:component_signal_ablation}(b) evaluates the design of the scoring mechanism within CTS. Using LRS alone leads to a drop in relative performance, indicating that the lack of cross-frame awareness leads to redundant selections across frames. Using SCS alone performs reasonably on MVBench but degrades noticeably on VideoMME, suggesting that relying solely on a candidate token's complementarity to the subspace spanned by retained evidence may not sufficiently account for its frame-local representativeness. Combining LRS and SCS achieves a relative performance of 97.0\%, outperforming either signal alone and supporting the benefit of
integrating the two signals within CTS.

\vspace{-4pt}

\paragraph{Ablation study on Complementarity Measure.}

Pairwise criteria evaluate a candidate against each retained token independently and may therefore overlook information jointly represented by multiple retained tokens and overvalue isolated outliers. To better account for such set-level relationships, SCS evaluates candidate complementarity against the subspace spanned by the retained evidence. To assess this design, we replace SCS with a pairwise-distance criterion while keeping all other settings unchanged. As shown in Table~\ref{tab:budget_complementarity_ablation}(b), this replacement reduces the relative performance from 97.0\% to 95.9\%, a drop of 1.1 points. These results support the effectiveness of subspace-level modeling for estimating cross-frame complementarity.

\subsection{Efficiency Analysis}
As shown in Table~\ref{tab:efficiency}, compared with the uncompressed model, GleanVID reduces prefill, generation, and total latency by 44.7\%, 40.9\%, and 24.1\%, respectively, while reducing peak GPU memory by 13.0\% and improving throughput by 32.0\%. Meanwhile, it preserves a VideoMME score of 63.6, only 0.9 points below the uncompressed model. Compared with VidCom2 and V-CAST, GleanVID incurs slightly higher prefill latency due to its sequential subspace updates, while achieving comparable total latency and throughput with the highest VideoMME score. Additional efficiency analyses are provided in Appendix~\ref{app:additional_efficiency}.

\section{Conclusion \& Limitation}\label{sec:conclusion}

We introduced GleanVID, a training-free visual-token compression framework for VideoLLMs. GleanVID allocates the global token budget across frames according to temporal novelty and selects tokens by balancing local representativeness with subspace-level complementarity to previously retained evidence. This two-level design prioritizes informative frames while reducing
redundancy in the retained visual evidence. Across three VideoLLMs and four benchmarks, GleanVID preserves performance under varying compression settings while reducing latency and GPU memory. However, SCS's linear-subspace formulation may not fully capture nonlinear token relations. Future work may explore richer complementarity models and broader deployment settings.

\bibliography{2027_conference}
\bibliographystyle{2027_conference}

\appendix
\clearpage
\appendix
\begin{tcolorbox}[
  enhanced, breakable,
  colback=oursbg, colframe=black!25,
  boxrule=0.6pt, arc=3mm,
  left=6pt, right=6pt, top=4pt, bottom=4pt
]
\begin{center}
  {\Large\bfseries Supplementary Material}
\end{center}
\vspace{-2pt}
\textbf{This appendix is organized as follows:}
\begin{itemize}[leftmargin=1.5em, itemsep=2pt, topsep=3pt]
  \item \textbf{Appendix~\ref{app:benchmark_details}} details the evaluated benchmarks, including their video durations, task categories, and evaluation formats.
  \item \textbf{Appendix~\ref{app:implementation_details}} provides implementation details, including the efficient implementation of TNB and CTS, the complete algorithms, computational complexity, LLM-side FLOPs estimation, and the evaluation setup.
  \item \textbf{Appendix~\ref{app:fixed_budget}} evaluates GleanVID under a fixed LLM-side visual-token budget with extended input frames.
  \item \textbf{Appendix~\ref{app:additional_ablation}} presents further ablations and hyperparameter sensitivity analyses, including component replacement with V-CAST, debiasing rank, temporal history length, neighborhood size, LRS temperature, and adaptive allocation bounds.
  \item \textbf{Appendix~\ref{app:additional_efficiency}} extends the efficiency analysis to LLaVA-OneVision-7B.
  \item \textbf{Appendix~\ref{sec:cross_frame_complementarity}} analyzes cross-frame evidence complementarity via effective rank, residual energy, and coverage error.
  \item \textbf{Appendix~\ref{app:qualitative}} presents qualitative comparisons with competing compression methods.
\end{itemize}
\end{tcolorbox}

\section{Benchmark Details}
\label{app:benchmark_details}

We evaluate GleanVID on four video-understanding benchmarks covering
different video durations and temporal reasoning requirements.

\paragraph{MVBench.}
MVBench~\citep{DBLP:conf/cvpr/0002WH00LWX0L0024} contains 4,000 multiple-choice
question-answering instances across 20 temporally sensitive tasks,
with video clips primarily ranging from 5 to 35 seconds. It evaluates
diverse capabilities, including action understanding, motion
perception, event ordering, and temporal reasoning.

\paragraph{LongVideoBench.}
LongVideoBench~\citep{wu2024longvideobench} contains 3,763 videos and
6,678 human-annotated multiple-choice questions across 17 categories,
with video durations ranging from 8 seconds to 1 hour. It evaluates
long-context understanding through referring-reasoning tasks that
require retrieving and integrating evidence from relevant temporal
contexts.

\paragraph{MLVU.}
MLVU~\citep{Zhou2024MLVUAC} contains 1,730 videos and 3,102 questions
across nine long-video understanding tasks, with video durations
ranging from approximately 3 minutes to more than 2 hours. Its tasks
evaluate holistic, single-detail, and multi-detail understanding
through both multiple-choice and free-form generation formats.

\paragraph{VideoMME.}
VideoMME~\citep{DBLP:conf/cvpr/FuDLLRZWZSZCLLZ25} contains 900 videos and 2,700
multiple-choice questions spanning six visual domains and 30
fine-grained categories. Its videos range from 11 seconds to 1 hour
and are evenly divided into short-, medium-, and long-video subsets,
enabling evaluation across different temporal scales.

\section{Implementation Details}
\label{app:implementation_details}

\subsection{Efficient implementation}
For Context Debiasing, we estimate the top-1 shared context direction
once from all centered visual tokens in a video and remove their
projections onto this direction. To efficiently compute the Temporal
Novelty in Eq.~\ref{eq:novelty}, let
$\mathbf{H}_{t,n}\in\mathbb{R}^{h_t\times D}$ denote the matrix
representation of the historical reference set
$\mathcal{H}_{t,n}$, whose rows are the historical neighborhood
prototypes associated with position $n$. We define
\begin{equation}
\begin{aligned}
\mathbf{G}_{t,n}
&=
\mathbf{H}_{t,n}\mathbf{H}_{t,n}^{\top}
+
\epsilon_{\mathrm{orth}}\,
\bar{d}_{t,n}\mathbf{I},
&
\bar{d}_{t,n}
&=
\frac{
\operatorname{tr}
(\mathbf{H}_{t,n}\mathbf{H}_{t,n}^{\top})
}{h_t},\\
\mathbf{b}_{t,n}
&=
\mathbf{H}_{t,n}\mathbf{z}_{t,n},
&
r_{t,n}
&=
\sqrt{
\left[
\|\mathbf{z}_{t,n}\|_2^2
-
\mathbf{b}_{t,n}^{\top}
\mathbf{G}_{t,n}^{-1}
\mathbf{b}_{t,n}
\right]_{+}
},
\end{aligned}
\label{eq:batched_gram}
\end{equation}
where $[a]_{+}=\max(a,0)$. The diagonal regularizer improves
numerical stability. We solve these Gram systems in parallel across
spatial locations, avoiding a separate QR decomposition for each token
trajectory.

For Complementary Token Selection, we incrementally maintain the
orthonormal basis $\mathbf{Q}_{t-1}$ spanning the debiased features of
previously selected tokens. Let $e_{t',n}^{(t-1)}$ denote the cached
squared SCS residual of an unprocessed token after processing the first
$t-1$ frames. After obtaining the new orthonormal directions
$\Delta\mathbf{Q}_{t}$ from frame $t$, we update
\begin{equation}
e_{t',n}^{(t)}
=
\left[
e_{t',n}^{(t-1)}
-
\left\|
\Delta\mathbf{Q}_{t}^{\top}
\widehat{\mathbf{x}}_{t',n}
\right\|_2^2
\right]_{+},
\qquad t'>t.
\label{eq:incremental_scs}
\end{equation}
Thus, the SCS used when processing frame $t'$ is
$v_{t',n}=\sqrt{e_{t',n}^{(t'-1)}}$. This incremental update
requires projections only onto the newly added directions, rather than
recomputing projections onto the complete subspace spanned by all
previously selected tokens.

Specifically, let $\mathbf{E}_{t}$ collect the residual components of
the tokens selected from frame $t$. We compute the reduced QR
decomposition
$\mathbf{E}_{t}^{\top}
=\Delta\mathbf{Q}_{t}\mathbf{R}_{t}$
and retain only directions satisfying
\begin{equation}
|(\mathbf{R}_{t})_{jj}|
>
\epsilon_{\mathrm{orth}}
\max_{\ell}|(\mathbf{R}_{t})_{\ell\ell}|.
\label{eq:qr_filter}
\end{equation}
This relative criterion filters numerically dependent directions
before $\Delta\mathbf{Q}_{t}$ is appended to
$\mathbf{Q}_{t-1}$.
\begin{algorithm}[t]
\caption{Temporal Novelty-Guided Budgeting}
\label{alg:tnb}
\begin{algorithmic}[1]
\Require Visual tokens
$\mathbf{X}=\{\mathbf{x}_{t,n}\}_{t=1,n=1}^{T,N}$;
global budget $B$; history length $h$;
allocation bounds $\alpha_{\min}$ and $\alpha_{\max}$
\Ensure Debiased features $\widehat{\mathbf{X}}$;
integer frame budgets $\{k_t\}_{t=1}^{T}$

\State Compute the global mean
$\bar{\mathbf{x}}
\gets (TN)^{-1}\sum_{t,n}\mathbf{x}_{t,n}$

\State Form the centered token matrix
$\mathbf{X}_{c}
\gets
[\mathbf{x}_{t,n}-\bar{\mathbf{x}}]_{t,n}$

\State Compute the top right singular vector
$\mathbf{v}_1$ of $\mathbf{X}_{c}$

\For{$t=1$ to $T$}
    \For{$n=1$ to $N$}
        \State $\widehat{\mathbf{x}}_{t,n}
        \gets
        (\mathbf{x}_{t,n}-\bar{\mathbf{x}})
        -
        \bigl((\mathbf{x}_{t,n}-\bar{\mathbf{x}})^\top
        \mathbf{v}_1\bigr)\mathbf{v}_1
        +
        \bar{\mathbf{x}}$
        \State $\mathbf{z}_{t,n}
        \gets
        |\mathcal{N}(n)|^{-1}
        \sum_{m\in\mathcal{N}(n)}
        \widehat{\mathbf{x}}_{t,m}$
    \EndFor
\EndFor

\State $r_{1,n}\gets\|\mathbf{z}_{1,n}\|_2$
for all $n$

\For{$t=2$ to $T$}
    \State $\mathcal{H}_{t,n}
    \gets
    \{\mathbf{z}_{t',n}\}_{t'=\max(1,t-h)}^{t-1}$
    for all $n$
    \State Compute
    $\{r_{t,n}\}_{n=1}^{N}$
    from $\{\mathcal{H}_{t,n}\}_{n=1}^{N}$
    using batched regularized Gram systems
\EndFor

\State $R_t\gets\sum_{n=1}^{N}r_{t,n}$
for all $t$

\State $s_t\gets R_t/\sum_{t'=1}^{T}R_{t'}$
for all $t$

\State $c\gets\operatorname{Gini}(R_1,\ldots,R_T)$

\State $\alpha
\gets
\alpha_{\max}
-
(\alpha_{\max}-\alpha_{\min})c$

\State $\widetilde{k}_t
\gets
\alpha B/T+(1-\alpha)B s_t$
for all $t$

\State $\{k_t\}_{t=1}^{T}
\gets
\Call{LargestRemainder}
{\{\widetilde{k}_t\}_{t=1}^{T},B,N}$

\State \Return
$\widehat{\mathbf{X}},\{k_t\}_{t=1}^{T}$
\end{algorithmic}
\end{algorithm}
\begin{algorithm}[t]
\caption{Complementary Token Selection}
\label{alg:cts}
\begin{algorithmic}[1]
\Require Raw features $\mathbf{X}$;
debiased features $\widehat{\mathbf{X}}$;
frame budgets $\{k_t\}_{t=1}^{T}$;
temperature $\tau$;
QR threshold $\epsilon_{\text{orth}}$
\Ensure Retained token index set $\mathcal{S}$

\State $\mathcal{S}\gets\varnothing$
\State $\mathbf{Q}\gets[\,]$
\Comment{Orthonormal basis of retained evidence}

\State $e_{t,n}\gets
\|\widehat{\mathbf{x}}_{t,n}\|_2^2$
for all $t,n$
\Comment{Cached residual energies}

\For{$t=1$ to $T$}
    \State Compute the frame prototype
    $\bar{\mathbf{f}}_t
    \gets
    N^{-1}\sum_{n=1}^{N}\mathbf{x}_{t,n}$

    \For{$n=1$ to $N$}
        \State Compute the local prototype
        $\bar{\mathbf{c}}_{t,\mathcal{N}(n)}
        \gets
        |\mathcal{N}(n)|^{-1}
        \sum_{m\in\mathcal{N}(n)}\mathbf{x}_{t,m}$

        \State $u_{t,n}
        \gets
        \tau^{-1}\operatorname{softplus}
        \left(
        \tau\left[
        \operatorname{sim}
        (\mathbf{x}_{t,n},
        \bar{\mathbf{c}}_{t,\mathcal{N}(n)})
        -
        \operatorname{sim}
        (\mathbf{x}_{t,n},\bar{\mathbf{f}}_t)
        \right]
        \right)$

        \State $v_{t,n}\gets\sqrt{\max(e_{t,n},0)}$
    \EndFor

    \State Normalize
    $\{u_{t,n}\}_{n=1}^{N}$
    and
    $\{v_{t,n}\}_{n=1}^{N}$
    independently to $[0,1]$,
    obtaining $\hat{u}_{t,n}$ and $\hat{v}_{t,n}$

    \State $g_{t,n}\gets\hat{u}_{t,n}+\hat{v}_{t,n}$
    for all $n$

    \State $\mathcal{T}_t
    \gets
    \operatorname{TopK}
    (\{g_{t,n}\}_{n=1}^{N},k_t)$

    \State $\mathcal{S}\gets\mathcal{S}\cup\mathcal{T}_t$

    \State Form the selected residual matrix
    $\mathbf{E}_t
    \gets
    [(\mathbf{I}-\mathbf{Q}\mathbf{Q}^{\top})
    \widehat{\mathbf{x}}_{t,n}]_{n\in\mathcal{T}_t}$

    \State Compute the reduced QR decomposition
    $\mathbf{E}_t^\top=\Delta\mathbf{Q}\mathbf{R}$

    \State Retain columns $\Delta\mathbf{q}_j$ satisfying
    $|R_{jj}|>
    \epsilon_{\text{orth}}\max_{\ell}|R_{\ell\ell}|$

    \State Append the retained directions:
    $\mathbf{Q}\gets[\mathbf{Q},\Delta\mathbf{Q}]$

    \For{$t'=t+1$ to $T$}
        \For{$n=1$ to $N$}
            \State $e_{t',n}
            \gets
            \max\left(
            e_{t',n}
            -
            \|\Delta\mathbf{Q}^{\top}
            \widehat{\mathbf{x}}_{t',n}\|_2^2,
            0
            \right)$
        \EndFor
    \EndFor
\EndFor

\State Sort $\mathcal{S}$ by the original token order
\State \Return $\mathcal{S}$
\end{algorithmic}
\end{algorithm}

Algorithms~\ref{alg:tnb} and~\ref{alg:cts} summarize the complete
procedures of Temporal Novelty-Guided Budgeting and Complementary Token
Selection, respectively.

\subsection{Computational Complexity}
\label{app:complexity}

Let $T$, $N$, and $D$ denote the number of frames, the number of
visual tokens per frame, and the token dimension, respectively. We
further denote the maximum temporal history length by $h$, the token
budget of frame $t$ by $k_t$, and the final rank of the accumulated
orthonormal basis by $q\leq\min(B,D)$, where
$B=\sum_{t=1}^{T}k_t$.

\paragraph{Compression complexity.}
Computing temporal novelty through the batched Gram systems requires
$\mathcal{O}(TN(h^2D+h^3))$ operations. Incremental SCS updates require
$\mathcal{O}(TNDq)$ operations, while the reduced QR decompositions
introduce $\mathcal{O}(D\sum_{t=1}^{T}k_t^2)$ additional cost.
Therefore, the overall computational complexity of GleanVID is
\begin{equation}
\mathcal{O}\!\left(
TN(h^2D+h^3+Dq)
+
D\sum_{t=1}^{T}k_t^2
\right).
\label{eq:overall_complexity}
\end{equation}
The incremental implementation avoids reconstructing the basis from
all previously selected tokens at every frame.

\paragraph{LLM-side FLOPs.}
The language backbones of the evaluated VideoLLMs employ grouped-query
attention and SwiGLU feed-forward networks. Following prior work~\citep{fan2026flashvid}, we estimate their prefill FLOPs as
\begin{equation}
F_{\mathrm{LLM}}(n)
=
L\left[
2nd^2\left(1+\frac{g}{a}\right)
+
2n^2d
+
3ndm
\right],
\label{eq:llm_flops}
\end{equation}
where $L$ is the number of Transformer layers, $n$ is the complete LLM
input length, $d$ and $m$ are the hidden and intermediate dimensions,
and $a$ and $g$ denote the numbers of query and key/value heads,
respectively.

Let $n_{\mathrm{full}}$ and $n_{\mathrm{comp}}$ denote the actual LLM
input lengths before and after compression, including visual, textual,
and special tokens. The resulting reduction in LLM-side prefill FLOPs
is
\begin{equation}
1-
\frac{
F_{\mathrm{LLM}}(n_{\mathrm{comp}})
}{
F_{\mathrm{LLM}}(n_{\mathrm{full}})
}.
\label{eq:flops_reduction}
\end{equation}
Since visual tokens constitute the majority of the input sequence,
reducing their number decreases both the quadratic attention cost and
the token-wise linear projection and FFN costs. We compute
Eq.~\ref{eq:llm_flops} using the architecture parameters of each
model and the actual sequence lengths observed before and after
compression.

\subsection{Evaluation setup}
\label{app:setup}
We evaluate GleanVID on Qwen3-VL-8B-Instruct, LLaVA-OneVision-7B, and LLaVA-Video-7B. Unless otherwise specified, all experiments are conducted on NVIDIA H800-80G GPUs and evaluated using the official LMMs-Eval toolkit. 
We use a shared default GleanVID configuration across all models and benchmarks unless stated otherwise. Specifically, we set the Context Debiasing rank to $r_d=1$ and use all available preceding frames for temporal novelty estimation. We use $2\times2$ neighborhoods to construct both the temporal prototypes in TNB and the local prototypes in LRS. The adaptive allocation bounds are set to
$\alpha_{\min}=0.2$ and $\alpha_{\max}=0.6$, and the LRS temperature is set to $\tau=4$. The relative tolerance $\epsilon_{\text{orth}}$ in Eq.~\ref{eq:batched_gram} and~\ref{eq:qr_filter} is set to $10^{-4}$.

\section{Additional Experimental Analysis}
\label{app:additional_experiments}


\providecommand{\methodtag}[1]{\,\mbox{\scriptsize\textnormal{(#1)}}}
\providecolor{oursbg}{RGB}{250,244,246}

\begin{table*}[t]
    \centering
    \captionsetup{position=top,skip=3pt}
    \caption{
        Performance comparison under a fixed LLM-side visual-token budget
        on Qwen3-VL-8B-Instruct. The uncompressed 16-frame baseline,
        80-frame inputs at 20\% retention, and 160-frame inputs at 10\%
        retention contain the same number of visual tokens after compression.
    }
    \label{tab:fixed_token_budget}

    \setlength{\tabcolsep}{4.5pt}
    \renewcommand{\arraystretch}{1.10}

    \resizebox{\textwidth}{!}{%
    \begin{tabular}{lcccccccccc}
        \toprule
        \multirow[c]{2}{*}{\textbf{Method}} &
        \multirow[c]{2}{*}{\textbf{\#Frames}} &
        \multirow[c]{2}{*}{%
            \shortstack{\textbf{Retention}\\\textbf{Ratio} $\rho$}} &
        \multirow[c]{2}{*}{%
            \shortstack{\textbf{LongVideo}\\\textbf{Bench}}} &
        \multirow[c]{2}{*}{\textbf{MLVU}} &
        \multicolumn{4}{c}{\textbf{VideoMME}} &
        \multicolumn{2}{c}{\textbf{Average}} \\
        \cmidrule(lr){6-9}
        \cmidrule(lr){10-11}
        & & & & &
        \textbf{Overall} &
        \textbf{Short} &
        \textbf{Medium} &
        \textbf{Long} &
        \textbf{Score} &
        \textbf{\%} \\
        \midrule
        \rowcolor{gray!12}
        \textbf{Vanilla} &
        16 (1$\times$) &
        100\% &
        57.1 &
        58.7 &
        60.6 &
        70.0 &
        57.8 &
        54.0 &
        58.8 &
        100.0 \\
        \midrule

        V-CAST\methodtag{2026'03} &
        &
        &
        61.8 &
        65.9 &
        66.1 &
        \textbf{77.7} &
        64.1 &
        56.4 &
        64.6 &
        109.9 \\

        \rowcolor{oursbg}
        \textbf{GleanVID}\methodtag{Ours} &
        \multirow{-2}{*}{80 (5$\times$)} &
        \multirow{-2}{*}{20\%} &
        \textbf{61.9} &
        \textbf{66.6} &
        \textbf{66.3} &
        \textbf{77.7} &
        \textbf{64.2} &
        \textbf{56.9} &
        \textbf{64.9} &
        \textbf{110.4} \\
        \midrule

        V-CAST\methodtag{2026'03} &
        &
        &
        61.0 &
        67.3 &
        65.9 &
        \textbf{78.0} &
        64.4 &
        55.3 &
        64.7 &
        110.1 \\

        \rowcolor{oursbg}
        \textbf{GleanVID}\methodtag{Ours} &
        \multirow{-2}{*}{160 (10$\times$)} &
        \multirow{-2}{*}{10\%} &
        \textbf{62.2} &
        \textbf{68.8} &
        \textbf{67.2} &
        77.2 &
        \textbf{66.6} &
        \textbf{57.8} &
        \textbf{66.1} &
        \textbf{112.4} \\
        \bottomrule
    \end{tabular}%
    }
\end{table*}

\subsection{Results under a Fixed Token Budget}
\label{app:fixed_budget}

To evaluate GleanVID's ability to leverage broader temporal evidence under a fixed LLM-side visual-token budget, we increase the number of input frames while proportionally reducing the retention ratio. Specifically, the uncompressed 16-frame baseline, 80-frame inputs at a 20\% retention ratio, and 160-frame inputs at a 10\% retention ratio contain the same number of visual tokens after compression. As shown in Table~\ref{tab:fixed_token_budget}, GleanVID achieves 110.4\% and 112.4\% relative performance with 80 and 160 input frames, respectively, outperforming V-CAST (109.9\% and 110.1\%) in both settings. The margin grows from 0.5 to 2.3 points as the input length increases and compression becomes more aggressive. These results show that GleanVID effectively converts token compression into broader temporal evidence coverage, with a clearer advantage under longer inputs and more aggressive compression.

\subsection{Additional Ablation and Sensitivity Analysis}
\label{app:additional_ablation}
\begin{table}[t]
\centering
\caption{Ablation study on component replacement on Qwen3-VL-8B-Instruct with 32 input frames and a 15\% retention ratio. Each variant replaces one GleanVID component with its V-CAST counterpart; V-CAST corresponds to replacing both.}
\label{tab:component_replace}
\small
\begin{tabular}{lccccc}
\toprule
\multirow{2}{*}{Variant} & \multirow{2}{*}{LongVideoBench} & \multicolumn{4}{c}{VideoMME} \\
\cmidrule(lr){3-6}
 & & Overall & Short & Medium & Long \\
\midrule
V-CAST & 57.6 & 62.0 & 71.9 & 58.3 & \textbf{55.9} \\
TNB $\rightarrow$ V-CAST budgeting & 58.0 & 62.1 & 72.3 & 58.9 & 55.1 \\
CTS $\rightarrow$ V-CAST scoring & 58.1 & 61.8 & 72.6 & 58.4 & 54.3 \\
\rowcolor{oursbg}
GleanVID & \textbf{59.0} & \textbf{62.6} & \textbf{73.6} & \textbf{59.6} & 54.6 \\
\bottomrule
\end{tabular}
\end{table}

\textbf{Ablation study on Component Replacement.}
To further assess whether each component improves upon a strong existing design, we replace TNB with the budget allocation of V-CAST and CTS with its token-scoring strategy, respectively, while keeping the other component unchanged. As shown in Table~\ref{tab:component_replace}, both replacements consistently degrade performance, reducing LongVideoBench by up to 1.0 point and VideoMME by up to 0.8 points, and replacing both components, which recovers V-CAST, yields the lowest scores. This suggests that TNB's content-level novelty estimation allocates budgets more effectively than frame-level trajectory cues, while CTS's subspace-level complementarity retains more informative evidence than token-wise scoring, with the two components providing complementary benefits.

\begin{table*}[t]
\centering

\begin{minipage}[t]{0.49\textwidth}
\vspace{0pt}
\centering
\captionof{table}{
Sensitivity analysis of the Context Debiasing rank on
Qwen3-VL-8B-Instruct with 32 input frames and a 15\% retention
ratio. A dagger ($\dagger$) marks the default setting.
}
\label{tab:debias_rank_sensitivity}
\vspace{2pt}

\setlength{\tabcolsep}{4.0pt}
\renewcommand{\arraystretch}{1.08}
\resizebox{\linewidth}{!}{
\begin{tabular}{lccccc}
\toprule
\multirow{2}{*}{Rank $r_d$}
& \multirow{2}{*}{LongVideoBench}
& \multicolumn{4}{c}{VideoMME} \\
\cmidrule(lr){3-6}
& & Overall & Short & Medium & Long \\
\midrule
$0$         & 58.6 & 62.4 & 73.2 & 59.1 & \textbf{54.8} \\
$1^\dagger$ & \textbf{59.0} & \textbf{62.6} & \textbf{73.6}
            & 59.6 & 54.6 \\
$2$         & 58.8 & 62.5 & 73.3 & 59.4 & \textbf{54.8} \\
$4$         & 58.8 & \textbf{62.6} & \textbf{73.6}
            & \textbf{60.0} & 54.1 \\
\bottomrule
\end{tabular}
}
\end{minipage}
\hfill
\begin{minipage}[t]{0.49\textwidth}
\vspace{0pt}
\centering
\captionof{table}{
Sensitivity analysis of the temporal history length on
Qwen3-VL-8B-Instruct with 32 input frames and a 15\% retention
ratio. A dagger ($\dagger$) marks the default setting.
}
\label{tab:history_length_sensitivity}
\vspace{2pt}

\setlength{\tabcolsep}{4.0pt}
\renewcommand{\arraystretch}{1.08}
\resizebox{\linewidth}{!}{
\begin{tabular}{lccccc}
\toprule
\multirow{2}{*}{History $h$}
& \multirow{2}{*}{LongVideoBench}
& \multicolumn{4}{c}{VideoMME} \\
\cmidrule(lr){3-6}
& & Overall & Short & Medium & Long \\
\midrule
$1$                    & 58.6 & 62.1 & 72.7 & \textbf{59.9} & 53.9 \\
$2$                    & 58.9 & 61.7 & 72.7 & 58.9 & 53.7 \\
$4$                    & 58.9 & 62.1 & 72.4 & 59.6 & 54.3 \\
$\mathrm{all}^\dagger$ & \textbf{59.0} & \textbf{62.6}
                       & \textbf{73.6} & 59.6 & \textbf{54.6} \\
\bottomrule
\end{tabular}
}
\end{minipage}

\end{table*}
\paragraph{Ablation study on Context Debiasing Rank.}
Context Debiasing is intended to reduce the influence of the shared
dominant feature direction on temporal novelty estimation.
Table~\ref{tab:debias_rank_sensitivity} evaluates the sensitivity of
GleanVID to its debiasing rank. Compared with disabling debiasing, the
default rank-one setting improves LongVideoBench from 58.6 to 59.0
and VideoMME Overall from 62.4 to 62.6. Increasing the rank to two or
four yields LongVideoBench scores of 58.8 and does not further improve
VideoMME Overall. These results suggest a possible trade-off: removing
the leading direction may help reduce shared variation, whereas
extending debiasing to additional directions provides no consistent
benefit and may also attenuate content-specific information. We
therefore use $r_d=1$ by default.
\paragraph{Ablation study on Temporal History Length.}
TNB estimates temporal novelty relative to previously observed
content, and the history length determines the temporal scope of this
reference. Table~\ref{tab:history_length_sensitivity} evaluates its
influence on frame-level budget allocation. Using all available
preceding frames achieves 59.0 on LongVideoBench and 62.6 on
VideoMME, compared with 58.6 and 62.1 when only the immediately
preceding frame is used. The settings with two and four historical
frames also remain below the full-history configuration on VideoMME
Overall. These results suggest that a broader history provides a more
complete reference for distinguishing current changes from previously
observed content, leading to more effective novelty-guided budget
allocation. We therefore use all available preceding frames by
default.

\begin{table*}[t]
\centering

\begin{minipage}[t]{0.49\textwidth}
\vspace{0pt}
\centering
\captionof{table}{
Sensitivity to the neighborhood size used for temporal prototypes in
TNB. A dagger ($\dagger$) marks the default setting.
}
\label{tab:temporal_neighborhood_sensitivity}
\vspace{2pt}

\setlength{\tabcolsep}{4.0pt}
\renewcommand{\arraystretch}{1.18}
\resizebox{\linewidth}{!}{
\begin{tabular}{lccccc}
\toprule
\multirow{2}{*}{Neighborhood}
& \multirow{2}{*}{LongVideoBench}
& \multicolumn{4}{c}{VideoMME} \\
\cmidrule(lr){3-6}
& & Overall & Short & Medium & Long \\
\midrule
$1\times1$         & 57.9 & 62.2 & 72.3 & 59.3 & \textbf{55.0} \\
$2\times2^\dagger$ & \textbf{59.0} & \textbf{62.6}
                   & \textbf{73.6} & \textbf{59.6} & 54.6 \\
$4\times4$         & 58.1 & 62.4 & 73.2 & \textbf{59.6} & 54.3 \\
\bottomrule
\end{tabular}
}
\end{minipage}
\hfill
\begin{minipage}[t]{0.49\textwidth}
\vspace{0pt}
\centering
\captionof{table}{
Sensitivity to the neighborhood size used for local prototypes in
LRS. A dagger ($\dagger$) marks the default setting.
}
\label{tab:lrs_neighborhood_sensitivity}
\vspace{2pt}

\setlength{\tabcolsep}{4.0pt}
\renewcommand{\arraystretch}{1.18}
\resizebox{\linewidth}{!}{
\begin{tabular}{lccccc}
\toprule
\multirow{2}{*}{Neighborhood}
& \multirow{2}{*}{LongVideoBench}
& \multicolumn{4}{c}{VideoMME} \\
\cmidrule(lr){3-6}
& & Overall & Short & Medium & Long \\
\midrule
$1\times1$         & 58.8 & 62.5 & 73.4 & \textbf{59.7} & 54.3 \\
$2\times2^\dagger$ & \textbf{59.0} & \textbf{62.6}
                   & \textbf{73.6} & 59.6 & \textbf{54.6} \\
$4\times4$         & 58.4 & 62.4 & 73.2 & 59.6 & 54.3 \\
\bottomrule
\end{tabular}
}
\end{minipage}

\end{table*}
\paragraph{Ablation study on Neighborhood Size.}
Both TNB and LRS construct local prototypes, but use them for different
purposes: TNB estimates temporal novelty, whereas LRS evaluates local
representativeness for token selection.
Tables~\ref{tab:temporal_neighborhood_sensitivity}
and~\ref{tab:lrs_neighborhood_sensitivity} examine their neighborhood
sizes by varying one setting at a time while keeping the other fixed.
For TNB, the default $2\times2$ neighborhood improves LongVideoBench
from 57.9 to 59.0 and VideoMME from 62.2 to 62.6 over pointwise
estimation. Enlarging it to $4\times4$ reduces LongVideoBench to 58.1
without further improving VideoMME. A similar trend is observed for
LRS: the $2\times2$ setting achieves the best aggregate results of
59.0 and 62.6, modestly outperforming pointwise estimation, whereas
the $4\times4$ setting degrades both scores. Despite minor
subset-level variations, these results consistently favor moderate
local aggregation: pointwise estimation may be sensitive to localized
variations, while a coarser neighborhood may obscure spatially
specific information. We therefore adopt a $2\times2$ neighborhood
for both TNB and LRS by default.


\begin{table*}[t]
\centering

\begin{minipage}[t]{0.49\textwidth}
\vspace{0pt}
\centering
\captionof{table}{
Sensitivity to the LRS temperature $\tau$.
A dagger ($\dagger$) marks the default setting.
}
\label{tab:lrs_temperature_sensitivity}
\vspace{2pt}

\setlength{\tabcolsep}{4.0pt}
\renewcommand{\arraystretch}{1.18}

\resizebox{\linewidth}{!}{%
\begin{tabular}{@{}cccccc@{}}
\toprule
\multirow{2}{*}{Temperature $\tau$}
& \multirow{2}{*}{LongVideoBench}
& \multicolumn{4}{c}{VideoMME} \\
\cmidrule(lr){3-6}
& & Overall & Short & Medium & Long \\
\midrule
$1$
& 58.3
& 61.6
& 72.3
& 58.7
& 53.8 \\

$2$
& 58.6
& 62.2
& 73.0
& 58.9
& 54.7 \\

$4^{\dagger}$
& \textbf{59.0}
& 62.6
& \textbf{73.6}
& 59.6
& 54.6 \\

$8$
& 58.6
& \textbf{62.8}
& \textbf{73.6}
& \textbf{59.9}
& \textbf{55.0} \\
\bottomrule
\end{tabular}%
}
\end{minipage}
\hfill
\begin{minipage}[t]{0.49\textwidth}
\vspace{0pt}
\centering
\captionof{table}{
Sensitivity to the adaptive allocation bounds.
A dagger ($\dagger$) marks the default setting.
}
\label{tab:allocation_bounds_sensitivity}
\vspace{2pt}

\setlength{\tabcolsep}{4.0pt}
\renewcommand{\arraystretch}{1.38}

\resizebox{\linewidth}{!}{%
\begin{tabular}{@{}cccccc@{}}
\toprule
\multirow{2}{*}{$[\alpha_{\min},\alpha_{\max}]$}
& \multirow{2}{*}{LongVideoBench}
& \multicolumn{4}{c}{VideoMME} \\
\cmidrule(lr){3-6}
& & Overall & Short & Medium & Long \\
\midrule
$[0.2,0.4]$
& 58.9
& \textbf{62.6}
& \textbf{73.7}
& 59.3
& \textbf{54.9} \\

$[0.2,0.6]^{\dagger}$
& \textbf{59.0}
& \textbf{62.6}
& 73.6
& 59.6
& 54.6 \\

$[0.2,0.8]$
& 58.7
& 62.4
& 73.2
& \textbf{59.8}
& 54.2 \\
\bottomrule
\end{tabular}%
}
\end{minipage}

\end{table*}
\paragraph{Ablation study on LRS Temperature.}
The temperature $\tau$ controls the softplus mapping of LRS values and
thereby affects token prioritization.
Table~\ref{tab:lrs_temperature_sensitivity} evaluates its effect.
Increasing $\tau$ from 1 to 4 improves LongVideoBench from 58.3 to
59.0 and VideoMME from 61.6 to 62.6. Further increasing $\tau$ to 8
raises VideoMME to 62.8 but reduces LongVideoBench to 58.6. These
results suggest that the temperature mediates a trade-off between
discriminability and stability in LRS-based token ranking. The default
$\tau=4$ may preserve sufficient separation between informative and
redundant tokens without making their relative priorities overly
sensitive to variations in LRS values. Increasing $\tau$ further shifts
this balance, resulting in benchmark-specific effects rather than
consistent improvements. We therefore use $\tau=4$ by default.

\paragraph{Ablation study on Adaptive Allocation Bounds.}
The allocation bounds determine the extent to which frame-level token
budgets can adapt to temporal novelty.
Table~\ref{tab:allocation_bounds_sensitivity} fixes
$\alpha_{\min}=0.2$ and varies $\alpha_{\max}$. The default range
$[0.2,0.6]$ achieves the highest LongVideoBench score of 59.0 and ties
for the best VideoMME score of 62.6. The alternative bounds exhibit
complementary duration-specific behavior: $\alpha_{\max}=0.4$ slightly
favors short and long videos, whereas $\alpha_{\max}=0.8$ favors
medium-length videos, but neither improves the aggregate results. This
pattern suggests that videos of different durations may benefit from
different degrees of allocation flexibility. Rather than optimizing
for a particular duration subset, $[0.2,0.6]$ maintains stronger
aggregate performance across both benchmarks, providing a suitable
balance between adaptive allocation and cross-duration consistency.
We therefore adopt it as the default range.

\subsection{Additional Efficiency Analysis}
\label{app:additional_efficiency}


\definecolor{effgreen}{RGB}{70,130,90}
\definecolor{dropred}{RGB}{170,70,70}
\definecolor{oursbg}{RGB}{234,239,244}
\renewcommand{\effdown}[1]{%
  {\scriptsize\textcolor{effgreen}{($\downarrow$#1\%)}}%
}

\newcommand{\perfgain}[1]{%
  {\scriptsize\textcolor{effgreen}{($\uparrow$#1)}}%
}

\renewcommand{\perfdrop}[1]{%
  {\scriptsize\textcolor{dropred}{($\downarrow$#1)}}%
}

\renewcommand{\methodtag}[1]{%
  {\scriptsize\textnormal{(#1)}}%
}

\begin{table*}[t]
\centering
\caption{
Efficiency and performance comparison on VideoMME using
LLaVA-OneVision-7B with 32 input frames and a retention ratio of
$\rho=25\%$. Efficiency reductions and performance changes are
computed relative to the uncompressed model.
}
\label{tab:llava_ov_efficiency}

\setlength{\tabcolsep}{8.0pt}
\renewcommand{\arraystretch}{1.12}

\resizebox{\textwidth}{!}{%
\begin{tabular}{lcccc}
\toprule
\textbf{Method}
& \multicolumn{1}{c}{
    \textbf{\shortstack{Prefill Latency $\downarrow$\\(s)}}}
& \multicolumn{1}{c}{
    \textbf{\shortstack{LLM Generation Latency $\downarrow$\\(s)}}}
& \multicolumn{1}{c}{
    \textbf{\shortstack{GPU Peak Memory $\downarrow$\\(MB)}}}
& \multicolumn{1}{c}{
    \textbf{Performance $\uparrow$}} \\
\midrule

\rowcolor{gray!10}
LLaVA-OneVision-7B
& 99.0
& 180.8
& 21,969.0
& 58.4 \\
\midrule

VidCom$^2$ \methodtag{EMNLP'25}
& \textbf{26.5}\effdown{73.2}
& \textbf{108.6}\effdown{39.9}
& \underline{19,512.6}\effdown{11.2}
& 58.4 \\

FastVID \methodtag{NeurIPS'25}
& 78.8\effdown{20.4}
& 160.4\effdown{11.3}
& 20,803.5\effdown{5.3}
& 58.3\perfdrop{0.1} \\

FlashVID \methodtag{ICLR'26}
& 28.4\effdown{71.3}
& 146.1\effdown{19.2}
& \textbf{19,059.9}\effdown{13.2}
& \underline{58.7}\perfgain{0.3} \\

V-CAST \methodtag{2026'03}
& 27.0\effdown{72.7}
& \underline{113.8}\effdown{37.0}
& \underline{19,512.6}\effdown{11.2}
& 58.2\perfdrop{0.2} \\

\rowcolor{oursbg}
\textbf{GleanVID} \methodtag{Ours}
& \underline{26.7}\effdown{73.1}
& 122.8\effdown{32.1}
& \underline{19,512.6}\effdown{11.2}
& \textbf{59.7}\perfgain{1.3} \\

\bottomrule
\end{tabular}%
}
\end{table*}
To examine whether the practical efficiency gains of GleanVID extend
beyond a specific model architecture, we further evaluate it on LLaVA-OneVision-7B. As shown in Table~\ref{tab:llava_ov_efficiency}, with 32 input frames and a retention ratio of $\rho=25\%$, GleanVID reduces prefill latency, LLM generation latency, and peak GPU memory by 73.1\%, 32.1\%, and 11.2\%,
respectively, relative to the uncompressed model. At the same time, it
improves the VideoMME score from 58.4 to 59.7. Among the evaluated compression methods, GleanVID offers a favorable joint
efficiency--performance profile: its prefill latency is only 0.2\,s above the minimum, while its VideoMME score is the highest. VidCom$^2$ and V-CAST reduce generation latency by an additional 14.2\,s and 9.0\,s, respectively, but trail GleanVID by 1.3 and 1.5 performance
points. GleanVID also achieves both lower generation latency and higher performance than FlashVID and FastVID. Overall, these results show that GleanVID substantially reduces inference cost without
compromising task performance, and that this trade-off remains consistent across different VideoLLM architectures.

\begin{figure*}[t]
    \centering
    \includegraphics[width=\textwidth]{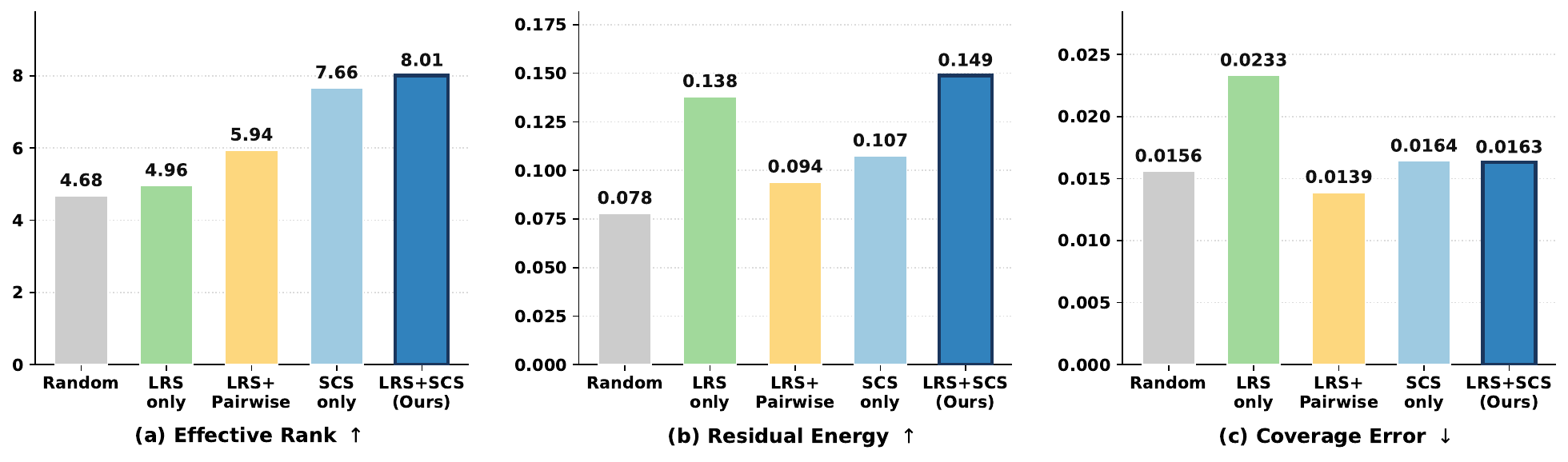}
    \caption{
    Cross-frame evidence complementarity analysis on 100
    LongVideoBench videos with 32 input frames and a 15\% retention
    ratio. All variants use identical per-frame token budgets.
    Higher Effective Rank and  Residual Energy indicate
    richer and more complementary retained evidence, while lower
    Coverage Error indicates better representation of discarded
    tokens.
    }
    \label{fig:cross_frame_complementarity}
\end{figure*}
\subsection{Cross-Frame Evidence Complementarity Analysis}
\label{sec:cross_frame_complementarity}

To assess whether the performance gains of GleanVID are consistent
with its intended mechanism of progressively accumulating
complementary evidence, we examine the geometric properties of the
retained token set. We consider three complementary metrics:
Effective Rank measures spectral diversity, Cross-Frame Residual
Energy quantifies the selected components unexplained by previously
retained evidence, and Coverage Error evaluates how well the final
retained subspace represents discarded tokens.

As shown in Figure~\ref{fig:cross_frame_complementarity}, the complete
LRS+SCS configuration achieves the highest Effective Rank of 8.01 and
Cross-Frame Residual Energy of 0.149. Compared with pairwise selection,
it improves these metrics by 34.8\% and 58.3\%, respectively,
indicating that subspace-aware selection retains a spectrally richer
set of evidence with larger components unexplained by the selection
history. Adding LRS to SCS further improves both metrics while
maintaining a nearly unchanged Coverage Error.
Although pairwise selection obtains the lowest Coverage Error, its
substantially lower Effective Rank and Residual Energy indicate that
average linear reconstruction does not necessarily imply richer
complementary evidence. Overall, LRS and SCS jointly provide a
favorable balance between local representativeness, cross-frame
complementarity, and global evidence coverage.

\section{Qualitative Analysis}
\label{app:qualitative}

\begin{figure*}[t]
    \centering
    \includegraphics[width=0.85\textwidth]{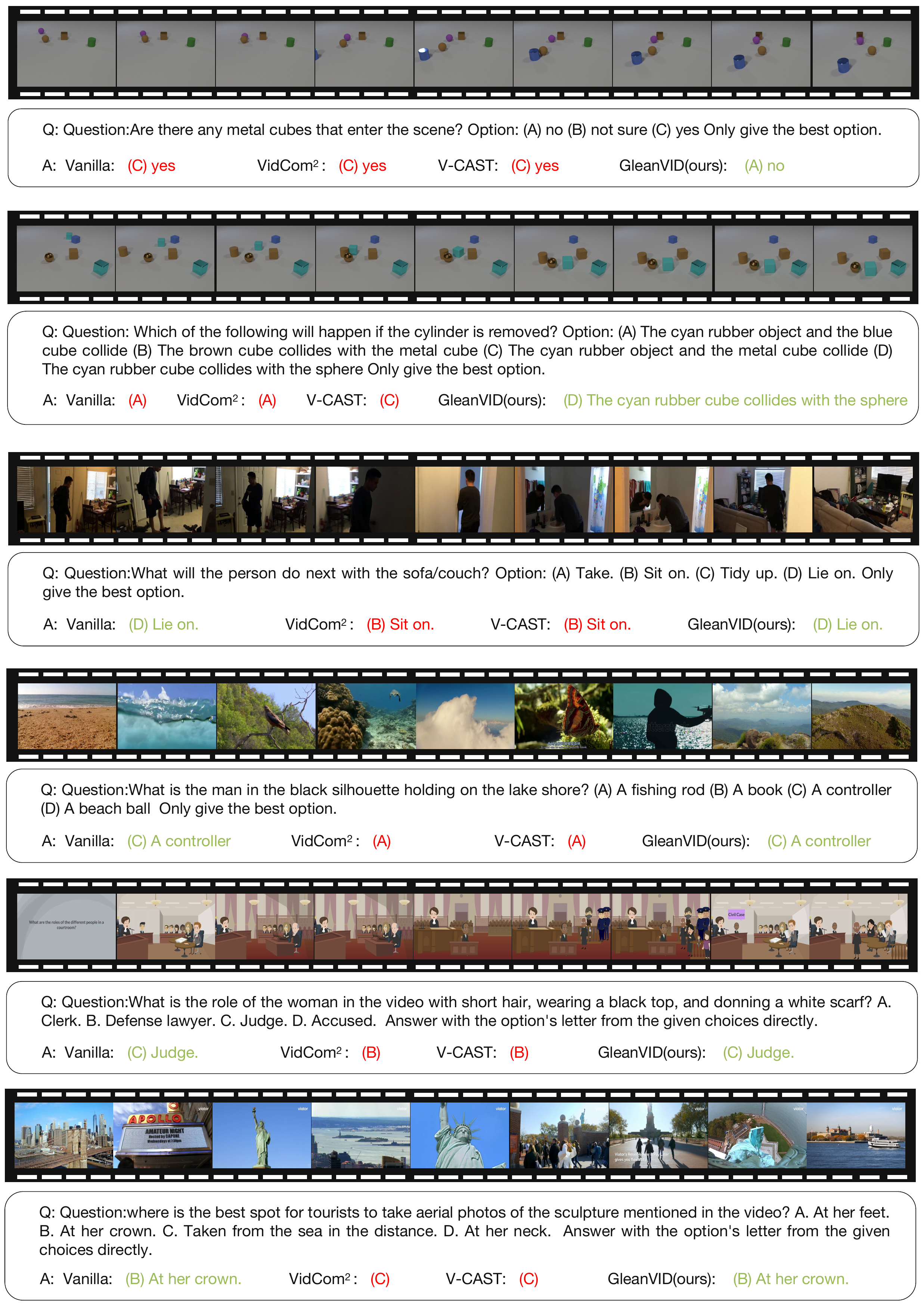}
    \caption{
    Qualitative comparison on LLaVA-OneVision-7B. Vanilla denotes the
    uncompressed model, while the remaining methods apply visual-token
    compression to the same model. The examples cover physical reasoning,
    action prediction, fine-grained grounding, and long-range video
    understanding. Correct and incorrect predictions are shown in green
    and red, respectively. GleanVID preserves task-relevant temporal and
    fine-grained evidence under a constrained visual-token budget.
    }
    \label{fig:qualitative_comparison}
\end{figure*}

Figure~\ref{fig:qualitative_comparison} presents qualitative
comparisons obtained by applying different token-compression methods to LLaVA-OneVision-7B, with Vanilla denoting the uncompressed model. The examples cover diverse video-understanding requirements, including object-state tracking, physical reasoning, action prediction, fine-grained visual grounding, and long-range semantic understanding. In the first two cases, GleanVID produces the correct answers even when the uncompressed model and competing compression methods fail. These examples suggest that selectively removing temporally redundant content may reduce its interference with downstream reasoning and help
the model focus on task-relevant visual evidence.

In the remaining cases, GleanVID retains the correct predictions of
the uncompressed model, whereas VidCom$^2$ and V-CAST lose temporal or
fine-grained cues required for action forecasting, object
identification, and long-range contextual reasoning. These observations
are consistent with the two-stage design of GleanVID: temporal
novelty-guided budgeting prioritizes frames containing newly introduced
content, while complementary token selection retains locally
representative evidence that is insufficiently covered by previous
selections. Overall, the qualitative results show that GleanVID
preserves task-relevant evidence while reducing cross-frame redundancy
under a constrained visual-token budget.

\end{document}